\documentclass[lettersize,journal,onecolumn]{IEEEtran}
\usepackage{amsmath,amsfonts}
\usepackage{algorithmic}
\usepackage{algorithm}
\usepackage{array}
\usepackage[caption=false]{subfig}
\usepackage{textcomp}
\usepackage{stfloats}
\usepackage{url}
\usepackage{verbatim}
\usepackage{graphicx}
\usepackage{cite}
\usepackage[utf8]{inputenc} 
\usepackage[T1]{fontenc}    
\usepackage{hyperref}       
\usepackage{url}            
\usepackage{booktabs}       
\usepackage{amssymb}
\usepackage{amsfonts}       
\usepackage{nicefrac}       
\usepackage{microtype}      
\usepackage{doi}
\usepackage{float}
\usepackage{xcolor}
\usepackage{amsmath, xparse}
\usepackage{orcidlink}
\usepackage{comment}

\title{Privacy-aware Berrut Approximated Coded Computing for Federated Learning}

\author{ Xavier
   Mart\'{\i}nez-Lua\~na$^\text{\orcidlink{0000-0001-9066-983X}}$, Rebeca
   P. D\'{\i}az-Redondo$^\text{\orcidlink{0000-0002-2367-2219}}$, Manuel
Fern\'andez-Veiga$^\text{\orcidlink{0000-0002-5088-0881}}$~\IEEEmembership{Senior
     Member,~IEEE} \thanks{X. Mart\'{\i}nez-Lua\~na is with the Galician
     Research and Development Center in Advanced Telecommunications
(GRADIANT)
     Estrada do Vilar, 56-58, Vigo, 36214, Spain. Email:
xmartinez@gradiant.org}
   \thanks{X. Mart\'{\i}nez-Lua\~na, R. P. D\'{\i}az-Redondo, and
     M. Fern\'andez-Veiga are with atlanTTic, Information \& Computing Lab,
     Telecommunication Engineering School, Universidade de Vigo Vigo, 36310,
     Spain. Emails: xamartinez@alumnado.uvigo.gal, rebeca@det.uvigo.es,
     mveiga@det.uvigo.es}
   \thanks{This work was partially funded by the European Union’s Horizon Europe
Framework Programme for Research and Innovation Action under project
TRUMPET (proj. no. 101070038), the grant PID2020-113795RB-C33
funded by MICIU/AEI/10.13039/501100011033 (COMPROMISE project),
the grant PID2023-148716OB-C31 funded by MCIU/AEI/ 10.13039/501100011033
(DISCOVERY project), "Contramedidas Inteligentes de Ciberseguridad para la
Red del Futuro (CICERO)" (CER-20231019) funded by CERVERA grants 2023 for RTOs,
partially funded by the European Union’s Horizon Europe Framework Programme for
Innovation Action under project TrustED (proj. no. 101168467) and ”TRUFFLES: TRUsted
Framework for Federated LEarning Systems", within the strategic cybersecurity
projects (INCIBE, Spain), funded by the Recovery, Transformation and Resilience
Plan (European Union, Next Generation). Additionally, it also has been funded
by the Galician Regional Government under project ED431B 2024/41 (GPC).

Funded by the European Union. Views and opinions expressed are however those of
the authors only and do not necessarily reflect those of the European Union.
Neither the European Union nor the granting authority can be held responsible
for them.
}  }

\hypersetup{
  pdftitle={Privacy-aware Berrut Aproximated Coded Computing for Federated Learning},
  pdfsubject={cs.IT, cs.NI},
  pdfauthor={X. Mart\'{\i}nez-Lua\~na},
  pdfkeywords={},
}

\begin{document}

\maketitle
\begin{abstract}
Federated Learning (FL) is an interesting strategy that enables the collaborative training of an AI model among different data owners without revealing their private datasets. Even so, FL has some privacy vulnerabilities that have been tried to be overcome by applying some techniques like Differential Privacy (DP), Homomorphic Encryption (HE), or Secure Multi-Party Computation (SMPC). However, these techniques have
some important drawbacks that might narrow their range of application: problems to work with non-linear functions and to operate large matrix multiplications and high communication and computational costs to manage semi-honest nodes. In this context, we propose a solution to guarantee privacy in FL schemes that simultaneously solves the previously mentioned problems. Our proposal is based on the Berrut Approximated Coded Computing, a technique from the Coded Distributed Computing paradigm, adapted to a Secret Sharing configuration, to provide input privacy to FL in a scalable way. It can be applied for computing non-linear functions and treats the special case of distributed matrix multiplication, a key primitive at the core of many automated learning tasks. Because of these characteristics, it could be applied in a wide range of FL scenarios, since it is independent of the machine learning models or aggregation algorithms used in the FL scheme. We provide analysis of the achieved privacy and complexity of our solution and, due to the extensive numerical results performed, a good trade-off between privacy and precision can be observed.
\end{abstract}


\section{Introduction}
\label{sec:introduction}


\IEEEPARstart{F}{ederated} Learning (FL)~\cite{FederatedLearning:KGN23} emerged as an adequate strategy to support collaborative training of AI models without sharing private data. The philosophy is not complex: (i) each worker node (or data owner) trains a local version of the AI model using its own private data; (ii) then, it shares the model (parameter, gradients, etc., depending on the AI algorithm) with a central computation node (or aggregator) which is in charge of (iii) assembling a global model, i.e. combining the partial knowledge of each node to obtain a global knowledge; and finally, (iv) sharing the global model with all the worker nodes in the collaborative (federated) network. This process is repeated until the global model converges. Despite its apparent simplicity, there are important challenges that arise in this new FL schemes, such as overheads in communication, management of non identically distributed computational resources and/or data, identification of suitable techniques to aggregate different models or privacy leaks because of the exchange of the AI models information.

In this paper, we exclusively focus on the privacy concerns that have been highlighted in the literature~\cite{lyu2022privacy, lyu2020threats}, such as the possibility of membership inference attacks~\cite{MIA:CCNS22, hu2022membership}, inferring class representatives~\cite{wang2019beyond}, inferring properties~\cite{melis2019exploiting} or even being able to reconstruct the original data by inverting gradients~\cite{geiping2020inverting}. These concerns have led to the emergence of different privacy enhancing technologies (PETs) focused on trying to avoid privacy leaks~\cite{yin2021comprehensive}, being the most used strategies the following ones:

\begin{enumerate}
    \item Differential privacy (DP)~\cite{DifferentialPrivacy:CD06} adds noise to the dataset before being locally used in the training stage, but it requires a high precision loss in order to achieve high privacy requirements.
    \item Homomorphic encryption (HE)~\cite{HomomorphicEncryption:CCD17} performs computations directly on the encrypted domain, but it requires high computation resources and it is difficult to apply with non-linear functions.
    \item Secure Multi-Party Computation (SMPC)~\cite{SecureMPC:EKR18}, closely related to the problem of secret sharing, allows a group of nodes to collaboratively compute a function over an input while keeping it secret. It entails high communication overheads and it cannot generally handle non-linear functions.
   
\end{enumerate}

However, these techniques have also some drawbacks. First, when dealing with semi-honest nodes\footnote{Semi-honest nodes participate in this kind of FL schemes using the information they receive to infer information of other parties. Since they do not interfere in the normal execution of the protocol, their detection is almost impossible.}, it is needed to add additional strategies to neutralize their behaviour, which increases the communication and computational costs and it would finally have a high impact in the scalability of the FL system~\cite{Kairouz2021}.

Besides, and since HE and SMPC are not suitable to deal with non-linear functions, the application of these techniques may narrow the range of AI models and aggregation algorithms used in FL schemes~\cite{sharma2017activation}. This would entail, first, usability implications, since there are AI models that cannot be applied in specific fields~\cite{George2014,Huda2016} and second, security implications, since some robust aggregation algorithms to deal with malicious nodes cannot be used~\cite{Pillutla2022,Li2023}. Additionally to the non-linear issues, managing matrix multiplications, which is needed in different AI algorithms, requires from high computation resources, aspect that has been addressed in the literature~\cite{Yuster2005,Smith2015,Zhang2020}.

Within this context, we propose to face the problem from the perspective of a large-scale distributed computing system, since FL naturally matches this scheme: a complex computation (global learning model) is distributedly performed by several worker nodes. Anyway, large-scale distributed computing systems have to deal with two important issues: communication overheads and straggling nodes. The former is due to the exchange of intermediate results in order to be able to collaboratively obtain the final one. The latter is due to slow worker nodes that reduce the runtime of the whole computation. Precisely, the Coded Distributed Computing (CDC) paradigm~\cite{Li20, CDCSurvey:NLLX20} arose to reduce both problems by combining coding techniques and redundant computation approaches. However, the CDC techniques cannot simultaneously guarantee privacy and work with non-linear functions. Besides, CDC schemes are usually applied for decentralized computations (one master node and one input dataset), but for FL (multiple datasets) it is more appropriate to work with a scheme like secret sharing protocols~\cite{beimel2011secret, chattopadhyay2024secret}, whose most famous approach is the Shamir Secret Sharing (SSS)~\cite{ShamirSecretSharing:S79}. In this context, we propose a CDC-based solution with the following contributions:

\begin{itemize}
\item [--] A CDC solution able to guarantee privacy and also able to manage non-linear functions. Recently, a CDC approach, the Berrut Approximated Coded Computing (BACC)~\cite{BACC:NA20}, was proposed to manage non-linear function in large distributed computation architectures. Thus, our first contribution defines a computation scheme that adds input privacy to the BACC algorithm, which we have coined as Privacy-aware Berrut Approximated Coded Computing (PBACC) scheme. In order to check the privacy-precision cost, we have performed a theoretical analysis of the privacy in PBACC.
\item [--] Extend the PBACC scheme to a multi-input secret sharing configuration, suitable for FL scenarios. Thus, our second contribution, coined as Private BACC Secret Sharing (PBSS), is based on the Shamir Secret Sharing, but assuring a secure FL aggregation process. We have also performed a theoretical analysis of the computational complexity of PBSS.
\item [--] Adapt the PBACC scheme to support optimized secure matrix multiplication. Due to the coding process, PBACC introduces additional computational load when multiplying matrices. Thus, our third contribution focuses on providing an appropriate multiplication mechanism that allows large matrix multiplication with three improvements: privacy-awareness, block partition and simultaneous encoding. We have also performed a theoretical analysis of the achieved privacy and a complexity analysis of our proposal.
\end{itemize}


In order to validate our proposal (PBSS), we have performed different experiments in FL settings to compare its behaviour in presence of stragglers with the original BACC with secret sharing without privacy. We have compared the precision of both schemes when dealing with non-linear functions and complex matrix multiplications. More specifically, we have performed experiments with three typical activation functions in machine learning algorithms (ReLU, Sigmoid and Swish) and two typical non-linear aggregation methods to combine learning models (Binary Step and Median)~\cite{sharma2017activation}. Besides, we have checked the performance of our proposal (PBSS) when operating complex matrices. In all cases the results are really promising, since the cost of adding privacy even in presence of stragglers is not significant in terms of precision.

The rest of the paper is organized as follows. Section~\ref{sec:related_work} briefly reviews some relevant works related to our proposal, and includes a description of the BACC approach~\cite{BACC:NA20}. Section~\ref{sec:pbacc} describes our first contribution PBACC and includes an analysis of the achieved privacy. In Section~\ref{sec:berrut-secret-sharing}, we detail how to extend PBACC to a multi-input secret sharing setting (PBSS), the analysis of its computational complexity, and a comparison with other SMPC and HE based approaches. In Section~\ref{sec:matrix-multiplication}, we detail our contribution to extend our scheme for matrix multiplication, including the privacy and complexity analysis, and a comparison with other secure matrix multiplication approaches. In Section~\ref{sec:results}, we test our proposal  and, finally, Section~\ref{sec:conclusions} summarizes the main conclusions of this work, and outlines some future work.

\section{Related Work}
\label{sec:related_work}

Coded distributed computing (CDC) techniques ~\cite{li2017fundamental, Li20, CDCSurvey:NLLX20} combine coding theory and distributed computing to alleviate the two main problems in large distributed computation. First, the high communication load due to the exchange of intermediate operations. This has been applied, for instance, in~\cite{lee2017speeding} for data shuffling, reducing communication bottlenecks by exploiting the excess in storage (redundancy). Second, the delay due to straggling nodes, which has been used in MapReduce schemes, for instance, to encode Map tasks even when not all the nodes had finished their computations~\cite{li2015coded}. 


Private Coded Distributed Computing (PCDC)~\cite{CDCSurvey2:UAGJ21} is a subset of Coded Distributed Computing (CDC) that focuses on preserving privacy of input data. Generally speaking, PCDC approaches try to compute a function among a set of distributed nodes but keeping the input data in secret. Thus, a central element divides the data into coded pieces that are shared with the computation nodes. Computation nodes, then, apply the goal function on these pieces of information (also known as {\em shares}). These partial results are, finally, sent to the central element to build the final result. The main research challenge in PCDC is trying to reduce the number of nodes that are required to carry out the computation of a given function.

One remarkable approach in the PCDC field is the Lagrange Coded Computing (LCC)~\cite{LCC:YLRK19}, which has been proposed as a unified solution for computing general multivariate polynomials (goal function) by using the Lagrange interpolation polynomial to create computation redundancy. LCC is resilient against stragglers, secure against malicious (Byzantine) nodes and adds privacy to the data set. Besides, and compare to other previous approaches, LCC reduces storage needs as well as reduces the communication and randomness overhead. However, LCC also has some important limitations: (i) it does not work with non-linear functions, (ii) it is numerically unstable when the input data are rational numbers or when the number of nodes is too large, and (iii) it relies on quantizing the input into the finite field. With the aim of overcoming the last of them, an extension of LCC for the analog domain was recently proposed~\cite{ALCC:SMA21}: Analog LCC (ALCC). However, it cannot solve the other issues related to the Lagrange interpolation. 

The Berrut Approximated Coded Computing (BACC)~\cite{BACC:NA20} offers a solution to the LCC or ALCC previously mentioned issues. BACC approximately computes arbitrary goal functions (not only polynomials) by dividing the calculus into an arbitrary large number of nodes (workers). The error of the approximation was theoretically proven to be bounded. However, BACC does not include any privacy guarantee.

All the previously mentioned CDC schemes are suitable for computations with one master node and one input data set. In FL settings, where there are multiple input data sets, one per node (worker), it would be more adequate to use the philosophy of the polynomial sharing approach~\cite{PolynomialSharing:NA21}, based on the Shamir Secret sharing (SSS)~\cite{ShamirSecretSharing:S79} approach. In SSS, the private information (secret) is split up in parts or shares, which are sent to the workers. The secret cannot be reassembled unless a sufficient number of nodes work together to reconstruct the original information.

Related to this last issue within SMPC, emerges the recurring problem of multiplying large matrices in a scalable and secure way. It is specially difficult to reduce communication and computation overheads (scalability) while maintaining privacy in the data and, besides, being resilient to stragglers. There are some approaches in the literature to face these issues, like ~\cite{PolynomialSharing:NA21}, which proposes a method for secure computation of massive matrix multiplication offloading the task to clusters of workers, or~\cite{KIM2023722}, a scheme for secure matrix multiplication in presence of colluding nodes. ~\cite{EntangledPolynomial:QA20} proposes entangled polynomial codes to break the cubic barrier of the recovery threshold for batch distributed matrix multiplications and~\cite{CodedSketch:21} proposes a coding scheme for batch distributed matrix multiplication resistant to stragglers.

\section{PBACC: Berrut Approximated Coded Computing with Privacy}
\label{sec:pbacc}

As previously mentioned, we have selected the Berrut Approximated Coded Computing (BACC)~\cite{BACC:NA20} as our building block for distributed computing because it works with non-linear computations, it is numerically stable, and has a lower computational complexity. However, BACC does not offer privacy guarantees, which is the issue we solved with our first contribution PBACC. In presence of colluding nodes, we define the privacy leakage of the scheme as the mutual information between the data and the messages captured  by the curious nodes. This mutual information can be upper bounded by the value of the Shannon capacity of an equivalent MIMO channel.

\subsection{Preliminaries}
\label{sec:preliminaries}

\begin{figure}[t]
    \centering
    \includegraphics[width=0.5\linewidth]{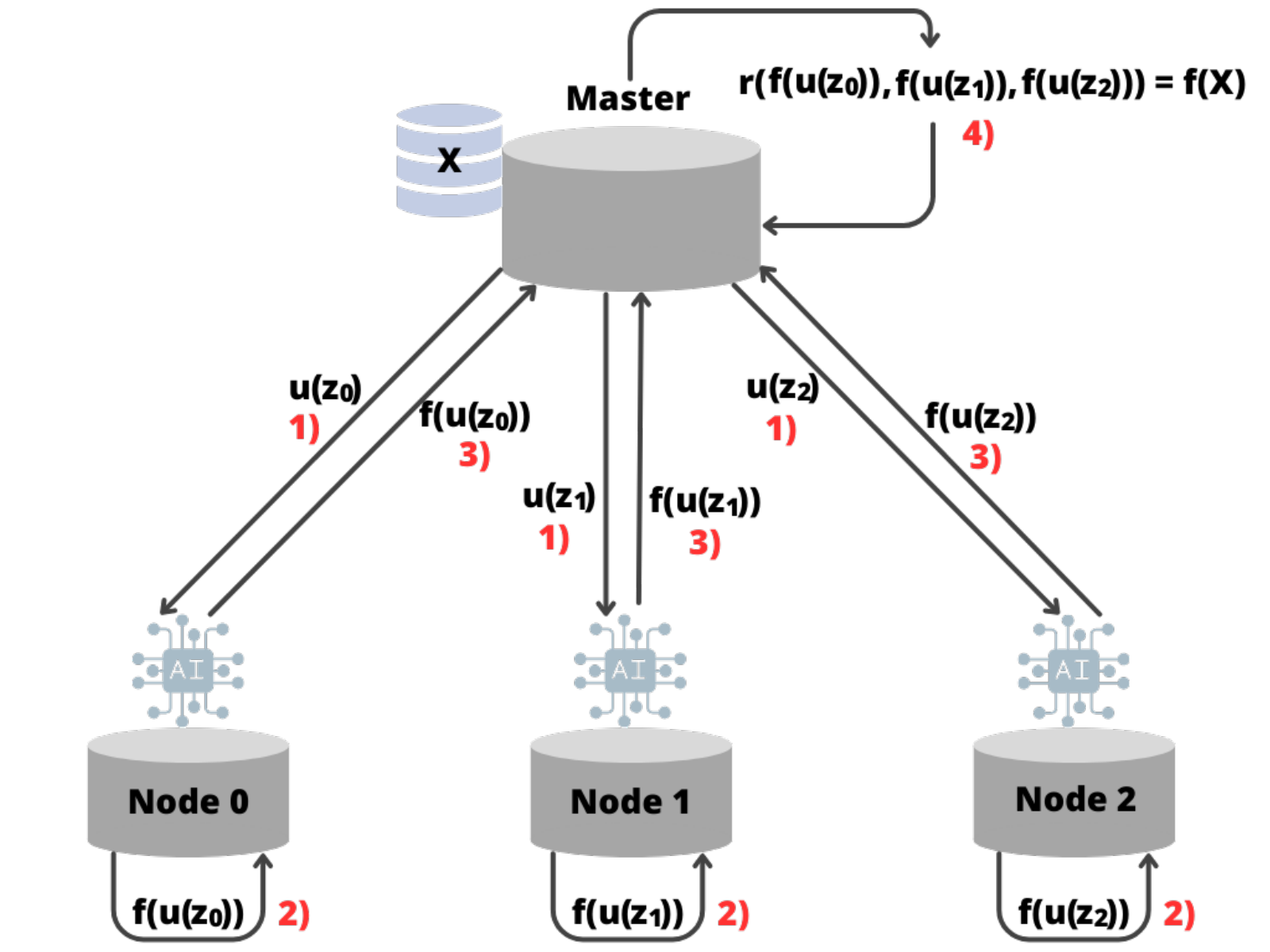}
    \caption{Decentralized computation of a function using BACC for 3 nodes.}
    \label{fig:decentralized_computation_bacc}
\end{figure}

Given an arbitrary matrix function $f: \mathbb{V} \rightarrow  \mathbb{U}$ over some input 
data  $\mathbf{X} = (X_0, \dots, X_{K-1})$, where $\mathbb{U}$ and $\mathbb{V}$ are vector 
spaces  of real  matrices, BACC performs the approximate evaluation of 
$f(\mathbf{X}) = \bigl( f(X_1), \dots, f(X_{K - 1})\bigr)$, where  $X_i\in \mathbb{V}$, in
a numerically stable form and with a bounded error. The computation is executed in a
decentralized configuration with one master node, who owns the data, and $N$ worker nodes
in charge of computing $f(\cdot)$ at some interpolation points which encode $\mathbf{X}$.
The general procedure is depicted in Fig.~\ref{fig:decentralized_computation_bacc}, and
explained next.

\subsubsection{Encoding and Sharing} To perform the encoding of the input data $\mathbf{X}$,
the following rational function $u: \mathbb{C} \rightarrow \mathbb{V}$ is composed by the 
master node
\begin{equation}
  \label{eq:bacc}
  u(z) = \sum_{i = 0}^{K - 1} \frac{\frac{(-1)^i}{(z - \alpha_i)}}{\sum_{j = 0}^{K - 1} 
  \frac{(-1)^j}{(z - \alpha_j)}} X_i, 
\end{equation}
for some distinct points $\boldsymbol\alpha = (\alpha_0, \dots, \alpha_{K-1}) 
\in \mathbb{R}^K$. By definition, it is  easy to  verify that 
$u(\alpha_j) = X_j$, for $j \in \{0, \dots, K - 1\}$. In the original BACC, it  is 
suggested to choose the interpolation points $\boldsymbol\alpha$ as the Chebyshev 
points of  first kind $\alpha_j = \operatorname{Re}(e^{\imath (2j + 1) \pi / (2K)})$
where $\imath = \sqrt{-1}$. Then, the master node selects $N$ points 
$\{ z_j: j = 0, \dots, N - 1 \}$, computes $u(z_j)$, and assigns this value to worker 
$j$. In~\cite{BACC:NA20}, it is suggested  to choose $\{ z_j: j = 0, \dots, N-1 \}$ 
as the Chebyshev points of second kind $z_j = \operatorname{Re}(e^{\imath j \pi / (N - 1)})$.

\subsubsection{Computation} Upon receiving $u(z_j)$, worker $j$ computes $v_j = f\bigl( u(z_j) \bigr)$.

\subsubsection{Communication} Worker $j$ sends the result $v_j$ to the master node.

\subsubsection{Decoding and recovery} When the master collects $n \leq N$ results from the 
subset $\mathcal{F}$ of fastest nodes, it approximately calculates $f(X_j)$, for 
$j = \{0, \dots, K -  1 \}$, using the decoding function based on the Berrut rational 
interpolation
\begin{equation*}
    r_{\mathrm{Berrut}, \mathcal{F}}(z) = \sum_{i = 0}^{n} \frac{\frac{(-1)^i}{(z - 
    \tilde{z}_i)}} {\sum_{j = 0}^{n}\frac{(-1)^j}{(z - \tilde{z}_j)}} f(u(\tilde{z}_i)),
\end{equation*} 
where $\tilde{z}_i \in \mathcal{S}$ are the evaluation points 
$\mathcal{S} = \{ \cos\frac{j\pi}{N-1}, j \in \mathcal{F} \}$ from the $n$ faster nodes. 
At this moment, the master node computes the approximation $f(X_i) \approx 
r_{\mathrm{Berrut}, \mathcal{F}}(\alpha_i)$, for $i \in \{0, \dots, K - 1 \}$.

\textbf{Threat Model}.
We assume that, in our system, from the $N$ total worker nodes,
up to $c$ nodes can be honest but curious, or semi-honest in short. This means
that these nodes, whose identities are unknown to the remaining honest nodes,
can attempt to disclose information on the private data, denoted by $\mathbf{X}$,
by means of observation of the encoded information received $\mathbf{Y}$ and direct
message exchange among them. Notice that semi-honest nodes respect the computation
protocol, i.e., they do not inject malicious or malformed messages to deceive the
master node.

Since we are dealing with rational functions, our scheme cannot achieve perfect
information-theoretical privacy, where the adversaries cannot learn anything about
the local data sets of honest clients. Instead, the focus is to achieve a bounded
information leakage lower than $\epsilon$, the target security parameter of the
scheme, which represents the maximum amount of leaked information per data point that
is allowed for a fixed number of colluding nodes $c$. The value of this parameter will
strongly depend on the specific ML model or application under consideration at each use
case.

\subsection{PBACC: Adding privacy to BACC}
\label{sec:adding_privacy_to_bacc}

As explained, the original BACC distributed computing solution is based on the rational
interpolation function~\eqref{eq:bacc} and the choice of $\boldsymbol\alpha$ as 
interpolation  points, so that  $u(\alpha_i) = X_i$ for all $i$. For the purpose of 
adding privacy, and similar to other works~\cite{Zhu2022,Raviv2019}, we introduce  
$T$ random coefficients into the encoding function, where $T$ is a design parameter. 
The encoding function is now
\begin{equation}
\label{eq:w(z)}
\begin{aligned}
    w(z) &= \sum_{i =0}^{K - 1} \frac{\frac{(-1)^{i}}{(z - \alpha_i)}}{\sum_{j = 0}^{K + T -1} 
    \frac{(-1)^j}{(z - \alpha_j)}} X_i + \sum_{i = 0}^{T - 1} \frac{\frac{(-1)^{K+i}}{(z - \alpha_{K + 
    i})}}{\sum_{j = 0}^{K + T - 1}\frac{(-1)^j}{(z - \alpha_j)}} R_i= \sum_{i = 0}^{K + T - 1} \frac{\frac{(-1)^{i}}{(z - \alpha_i)}}{\sum_{j = 0}^{K + T - 1}\frac{(-1)^j}{(z - 
    \alpha_j)}} W_i,
\end{aligned}
\end{equation}
where $\mathbf{W} = (X_0, \dots, X_{K-1}, R_0, \dots, R_{T-1}) = 
(W_0, \dots, W_{K + T - 1})$. Here, $\{R_{i}: i = 1, \dots, T - 1 \}$
are random data points independently generated according to a given distribution $\mathcal{G}$
to be specified later. To be able to accomplish the interpolation of the values of $f(X_i)$
for all $i$, we propose to select as interpolation points of $X$, the Chebyshev points
of first kind, and as interpolation points of $R$, the shifted Chebyshev points of first
kind
$\beta_{j} = b + \alpha_j = b + \operatorname{Re}(e^{\imath (2j + 1) \pi / (2T)})$ where
$b \in \mathbb{R}$, for $j = 0, \dots, T - 1$.

Re-writing eq.~\eqref{eq:w(z)} as 
\begin{equation*}
w(z) = \sum_{i=0}^{K-1} X_i q_i(z) + \sum_{i=K}^{K + T - 1} R_{i - K} q_i(z),
\end{equation*}
where $q_i(z)$ is:
\begin{equation*}
    q_i(z) = \frac{\frac{(-1)^i}{(z-\alpha_i)}}{\sum_{j=0}^{K+T-1}\frac{(-1)^j}{(z-\alpha_j)}},
\end{equation*}
we can easily check that $q_i(z)=1$ when $z=\alpha_i$, and $q_i(z)=0$ when $z=\alpha_j$ for all
$j \neq i$. This demonstrates the identity $w(\alpha_i) = X_i$ for $i = \{0, 1, \dots, K-1\}$,
and $w(\alpha_j) = R_j$ for $j = \{K, K+1, \dots, K+T-1\}$. Thus, the Berrut rational interpolation
can be used for interpolating the values of $X_i$.

The proposed scheme will be referred to as Privacy-aware Berrut Approximated Coded
Computing (PBACC) hereafter. 

In the way we have added privacy, PBACC inherits the features of the original BACC (dealing with non-linear
functions and resisting stragglers) at a very low cost on precission loss. Our experimentation on the scheme
(Section~\ref{sec:results}) demonstrates this behaviour.


\subsection{Measuring the privacy of PBACC}
\label{sec:measuring_privacy_of_pbacc}

As stated in the threat model, our focus is not on perfect information-theoretical
privacy but on bounded information leakage.
Therefore, similar to other works~\cite{ALCC:SMA21,Issa2020}, we use the  mutual information
$I(\mathbf{X}; \mathbf{Y})$, which is a measure of the statistical correlation of two random
variables $\mathbf{X}$ and $\mathbf{Y}$, as the privacy leakage metric. One of the variables
$\mathbf{X}$ will correspond to the private input, while the other $\mathbf{Y}$ will represent
the encoded output, and is a worst-case for the information accessible to the curious or
semi-honest colluding nodes. Specifically, we assume that, in our system, up to $c$ nodes
can be honest but curious, or semi-honest in short.

Generally, the exact computation of $I(\mathbf{X}; \mathbf{Y})$ under the described threat
model is not feasible. Instead, since our goal is to obtain a bounded leakage assurance, we
leverage the existent results about the capacity of a Multi-Input Multi-Output (MIMO)
channel under some specific constraints~\cite{MIMOCapacity:SPM02}, to bound the
mutual information, using similar techniques to~\cite{ALCC:SMA21}. Such bound is based 
in the fact that a  MIMO channel with $K$ transmitter antennas and $c$ receiver antennas
is, conceptually from an information-theoretic perspective, equivalent to a signal
composed of the $K$ encoded elements of the input in PBACC, and an observation
cooperatively formed by the $c$ colluding nodes, which collect and possibly process these
encoded messages to infer information about the data.
In this configuration, the encoded noise of  the channel acts as the element that provides
security, since it reduces the amount of  information received by the colluding nodes. We assume for the analysis that follows that the noise is Gaussian (so the channel is an AWGN
vector channel).

Now, for this privacy analysis, we start with the encoding function~\eqref{eq:w(z)}, 
written as $w(z) = \sum_{i=0}^{K-1} X_i q_i(z) + \sum_{i=K}^{K + T - 1} R_{i - K} q_i(z)$,
where
\begin{equation*}
    q_i(z) = \frac{\frac{(-1)^i}{(z-\alpha_i)}}{\sum_{j=0}^{K+T-1}\frac{(-1)^j}{(z-\alpha_j)}}
\end{equation*}
is one of the rational interpolating terms. Without loss of generality, we assume that
each entry of $X_i$ is a realization of a random variable with a distribution supported 
in  the interval $D_s \triangleq [-s, s]$, for all $i = \{0, \dots, K-1\}$, and that 
the  random coefficients $R_i  \sim \mathcal{N}(0, \frac{\sigma_n^2}{T})$, where 
$\mathcal{N}(0, a)$ denotes the Gaussian distribution with zero mean and variance $a$. 
Note that, having a finite support $D_s$, $X_i$ has finite power. If we denote 
$\mathbf{Y} = (w(z_{i_1}), w(z_{i_2}), \dots, w(z_{i_{c}}))^T$, for the messages observed 
by the $c$ colluding adversaries
$\mathcal{C} = \{ i_1, i_2, \dots, i_{c} \}$, 
where $z_{i_h}$ are some of the Chebyshev points of second kind, we can write
\begin{equation}
 \label{eq:vector-BACC}
 \mathbf{Y} = Q_{c} \mathbf{X} + \tilde{Q}_{c} \mathbf{R},
\end{equation}
where the matrices $Q_{c}$ and $\tilde{Q}_{c}$ are given by
\begin{equation*}
\begin{aligned}
 &Q_{c} \triangleq
    \begin{pmatrix}
        q_{0}(z_{i_1}) & \hdots & q_{K-1}(z_{i_1}) \\
        q_{0}(z_{i_2}) & \hdots & q_{K-1}(z_{i_2}) \\
        \vdots & \ddots & \vdots \\
        q_{0}(z_{i_{c}}) & \hdots & q_{K-1}(z_{i_{c}})
    \end{pmatrix}_{c \times K},\\
  &\tilde{Q}_{c} \triangleq 
    \begin{pmatrix}
        q_{K}(z_{i_1}) & \hdots & q_{K+T-1}(z_{i_1}) \\
        q_{K}(z_{i_2}) & \hdots & q_{K+T-1}(z_{i_2}) \\
        \vdots & \ddots & \vdots \\
        q_{K}(z_{i_{c}}) & \hdots & q_{K+T-1}(z_{i_{c}})
    \end{pmatrix}_{c \times T}.
\end{aligned}
\end{equation*}
The amount of information revealed to $\mathcal{C}$  ---the information leakage $I_L$--- 
is defined in this work as the worst-case achievable mutual information
for the colluding  nodes~\cite{Issa2020}
\begin{equation*}
    I_L \triangleq \max_{\mathcal{C}} \sup_{P_\mathbf{X}: ||X_i| \leq s|, \forall i \in [K]}  
    I(\mathbf{Y}; \mathbf{X}),
\end{equation*}
where $P_\mathbf{X}$ is the probability density function of $\mathbf{X}$, and the 
maximization applies to any $\mathcal{C} \subset [N]$. As $\|X_i\| \leq s$, this implies 
that the power $\mathbb{E}[ \|X_i\| ]^2 \leq s^2$. Therefore, the latter equation can 
be re-written as
\begin{equation}
  \label{eq:eta_c}
    I_L \leq \max_{\mathcal{C}} \sup_{P_\mathbf{X}: \mathbb{E}[ \|X_i \|^2] \leq s^2}  
    I(\mathbf{Y}; \mathbf{X}).
\end{equation}
In order to bound $I_L$, since the noise used for privacy in PBACC
is Gaussian, we consider an equivalent formulation of~\eqref{eq:vector-BACC} as a MIMO channel
with $K$ transmitter antennas and $c$ receiver antennas, so the
input-output relation can be defined as
\begin{equation*} 
  \mathbf{Y} = H \mathbf{X} + \mathbf{Z},
\end{equation*}
where $H_{c \times K}$ is the channel gain matrix known by both transmitter and receiver, 
and $\mathbf{Z}_{c \times 1}$ is a vector of additive Gaussian noise with 0 mean. Further,
let us denote the covariance matrix of the vector $\mathbf{Z}$ as $\Sigma_{\mathbf{Z}}$.
Using known results on the capacity of a MIMO channel with the same power allocation 
constraint and correlated noise~\cite{MIMOCapacity:SPM02},
\begin{equation*}
    C = \sup_{P_{\mathbf{X}}} I(\mathbf{Y}; \mathbf{X}) = \log_2 | I_{c} + P H 
    \Sigma^{-1}_{\mathbf{Z}} H^{\dagger} |,
\end{equation*}
where $P$ is the maximum power of each transmitter antenna, $I_{c}$ is the identity matrix 
of order $c$ and $|\cdot|$ is the determinant of a matrix. Thus, using~\eqref{eq:eta_c}
and assuming that the noise is uncorrelated, we have
\begin{equation*}\
  \label{eq:updated_eta_c}
    I_L \leq \max_{\mathcal{C}} \log_2 \bigl| I_{c} + \frac{s^2 T}{\sigma^2_n} \tilde{\Sigma}_{c}^{-1} \Sigma_{c} \bigr|,
\end{equation*}
where $\tilde{\Sigma}_{c} \triangleq \tilde{Q}_{c} \tilde{Q}_{c}^{\dagger}$ and
$\Sigma_{c} \triangleq Q_{c}
Q_{c}^{\dagger}$.

Since the information leakage grows with $K$, the privacy metric $\imath_L$ of
PBACC is defined in this paper as the  normalized value of
$I_L$, namely $\imath_L = \frac{I_L}{K}$, and we will say that
PBACC has $\epsilon$-privacy if $\imath_L < \epsilon$, where
$\epsilon > 0$ is the target security parameter of the scheme that represents the
maximum amount of leaked information per data point that is allowed for a fixed number
of colluding nodes $c$, as stated in the threat model.

To demonstrate that we meet the privacy assumptions specified in Section~\ref{sec:preliminaries},
we will analyze the evolution of the mutual information in function of the number of colluding nodes
for different values of $\sigma_n$. The analysis will show if our encoding provides privacy to $X$
and clarify if we can meet the privacy requirements specified in the threat model.

\begin{figure}[t]
    \centering
    \includegraphics[width=0.5\linewidth]{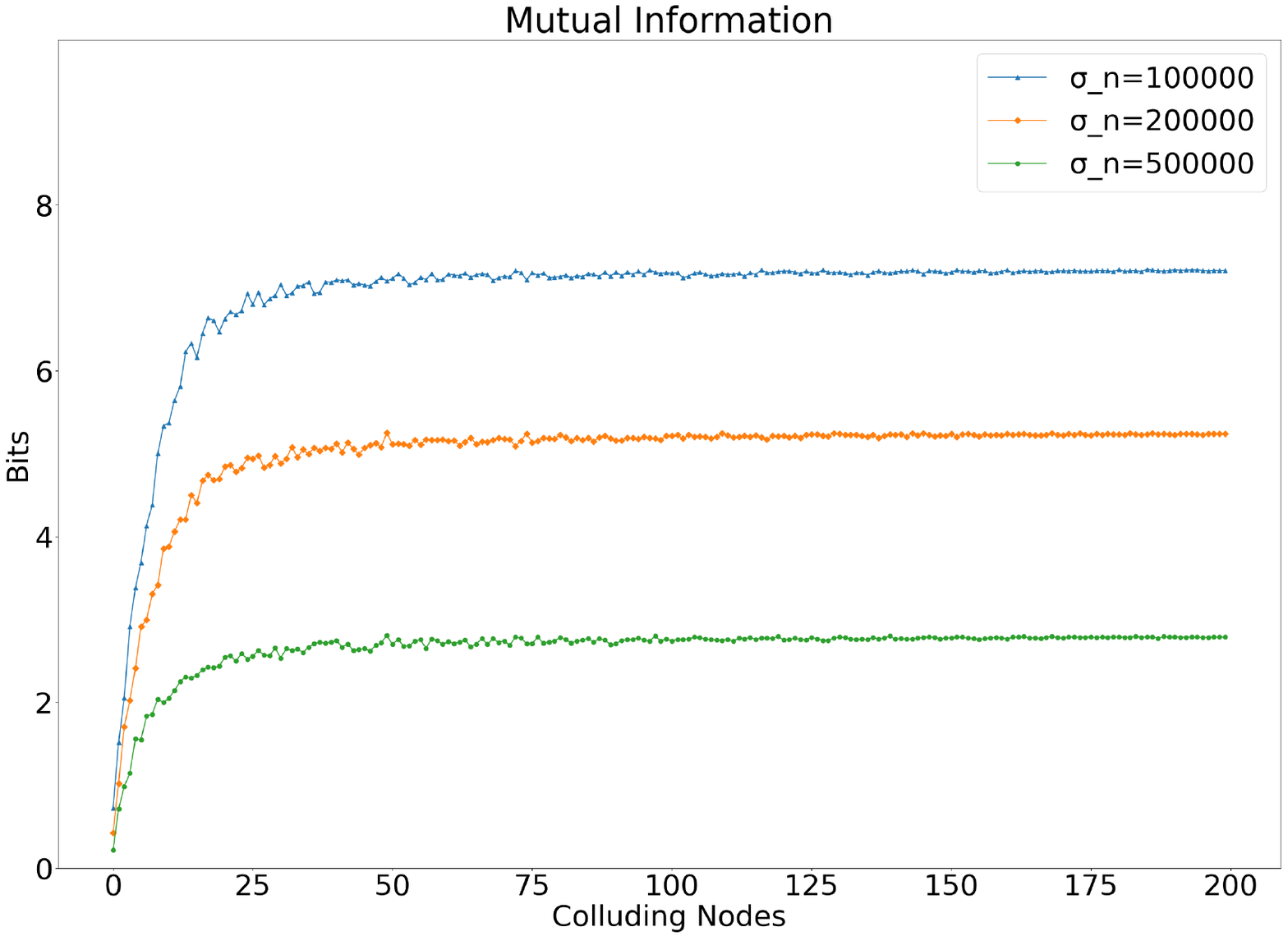}
    \caption{Evolution of $I_L$ for $T=50$, $\sigma_n=\{10^5, 2\times10^5, 5\times10^5\}$}
    \label{fig:privacy_evolution}
\end{figure}

We start by setting up a very pessimistic scenario where we assume that if all the $N=200$ nodes of the
network collude, in presence of some minimum noise $T=1$, and $s=\sigma_n$, they can recover all the
information of $X$, being $K = 10$ and $s=10^4$. This implies that $I_L = H(X)$ at most, being $H()$ the
entropy of a random variable. Since the covariance matrix $\tilde{\Sigma}_c$ has to be inverted and it is
ill-conditioned for this parameters, we regularize it using the Minimum Eigenvalue Method (very fitting for
regularizing covariance matrices as per \cite{Tabeart20}). We consider that $X_i \sim U(-s,s)$, where $U(-a, a)$ denotes the Uniform distribution in the interval $D_s\triangleq [-a, a]$, to maximize the entropy. Thus, we regularize the covariance matrix until $I_L \approx 14.28$. At this moment, we set $T=50$ and we test the evolution of
the mutual information in function of colluding nodes $c$ for
$\sigma_n=\{10^5, 2\times10^5, 5\times10^6\}$. Fig.~\ref{fig:privacy_evolution} shows the evolution of $I_L$,
and we can clearly see that, even after starting from such a pessimistic scenario, increasing the ammount of random
coefficients $T$ and the standard deviation of the noise $\sigma_n$ reduces the mutual information, and we
can control this value to meet an $\epsilon$-security value suitable for the specific use case in study.
Furthermore, we can check that the mutual information scales logarithmically, which means that new colluding
nodes of the network only infer information already known by the rest of the group, which is a valuable security
feature for this type of configurations.

\color{black} 




\section{PBSS: PBACC with Secret Sharing}
\label{sec:berrut-secret-sharing}

While PBACC provides quantifiable privacy for distributed computations, it is not yet 
valid for FL, since it only supports a single data owner (the master). The workers are just
delegated nodes for the computation tasks. In contrast, in FL the roles are reversed:
the clients perform local training and are also the data owners, whereas the aggregator
node is the central entity doing the model aggregation. Therefore, to be useful for general 
FL frameworks, PBACC needs to be enhanced to a multi-input secret sharing configuration, one
that allows a set of nodes to collaboratively compute a function, either training or 
aggregation, whose final results can be decoded by the central aggregator. After the 
protocol is defined, it is also essential to measure its communication and computation
costs in each phase, so that it proves to be scalable in addition
to privacy-preserving. In the following, the Private BACC Secret Sharing (PBSS) protocol is
presented (Section~\ref{sec:bss_ss_protocol}), and its computational complexity is analyzed
(Section \ref{sec:bss_scalability}). Finally, we have added a comparison of the
proposed scheme with some SMPC and HE approaches for non-linear function computations (\ref{sec:pbss_comparison}).
\color{black}

\subsection{Private BACC Secret Sharing (PBSS) protocol} 
\label{sec:bss_ss_protocol}

\begin{figure*}[t]
    \centering
    \subfloat[Phase 1 of PBSS- Sharing.\label{fig:bss_sharing_phase}]{%
        \includegraphics[width=0.49\linewidth]{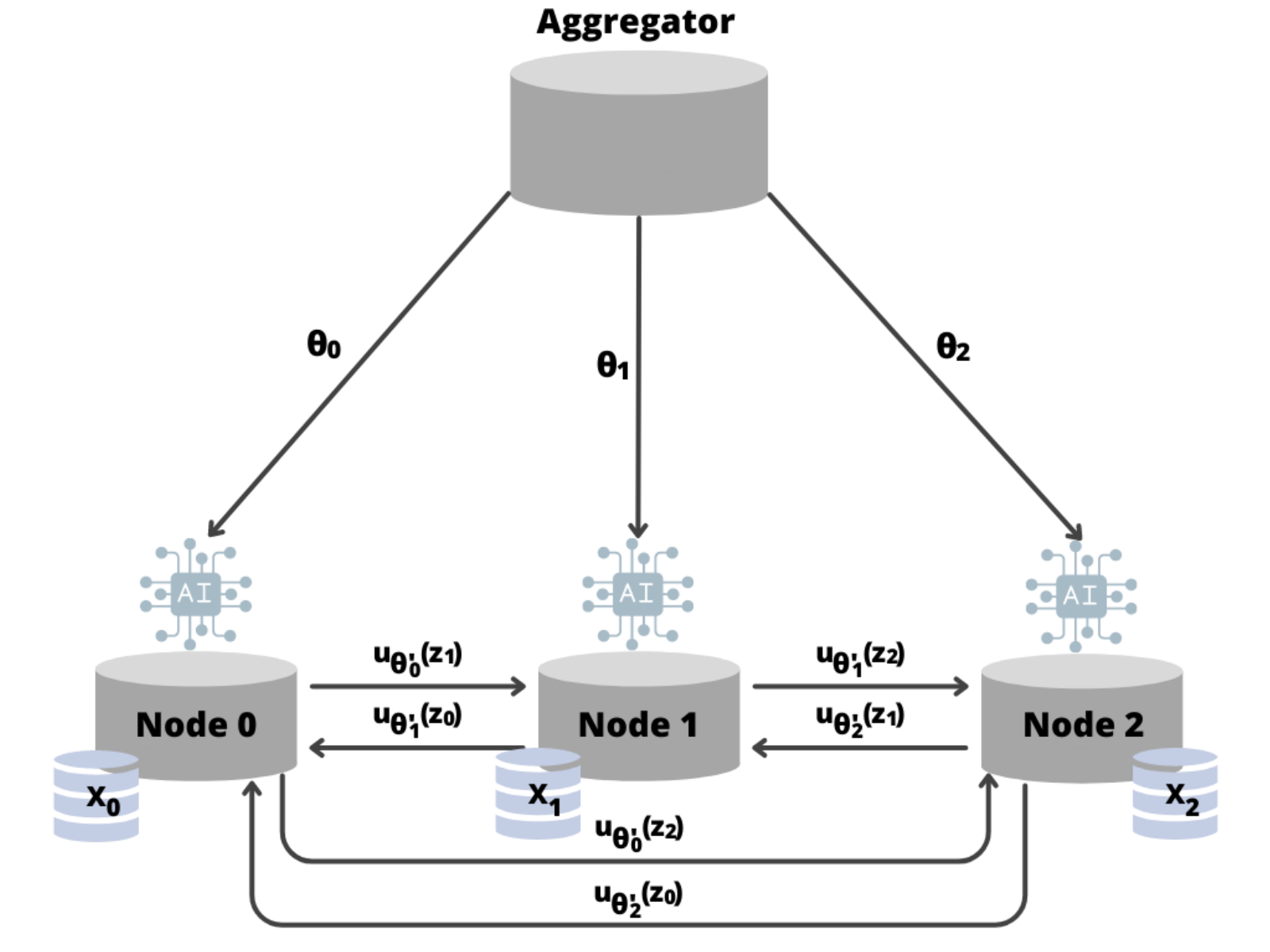}
    }
    \subfloat[Phase 2 of PBSS - Computation and communication.\label{fig:bss_computation_communication_phase}]{%
        \includegraphics[width=0.49\linewidth]{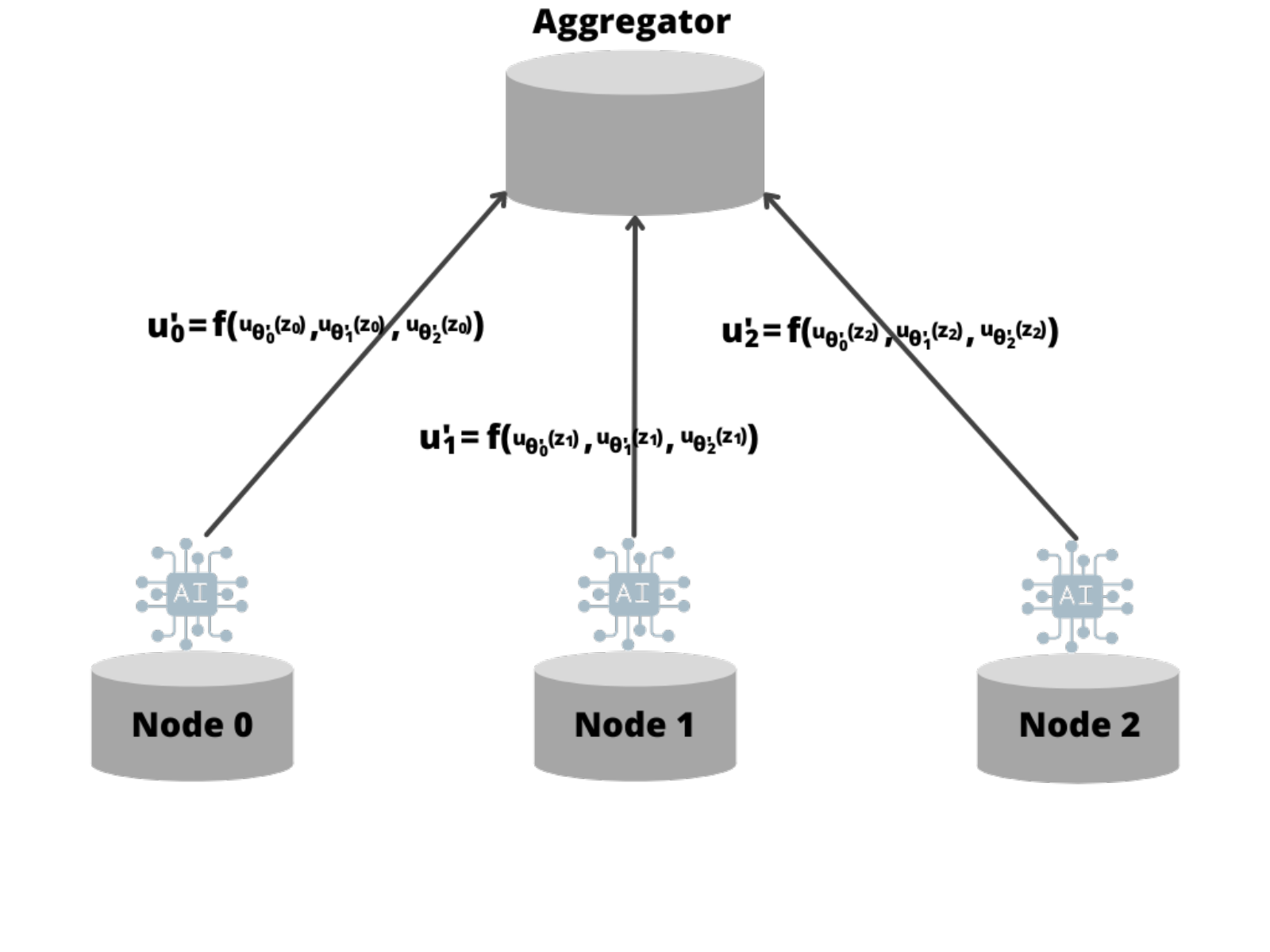}
    }\\
    \subfloat[Phase 3 of PBSS - Reconstruction phase.\label{fig:bss_reconstruction_phase}]{%
        \includegraphics[width=0.49\linewidth]{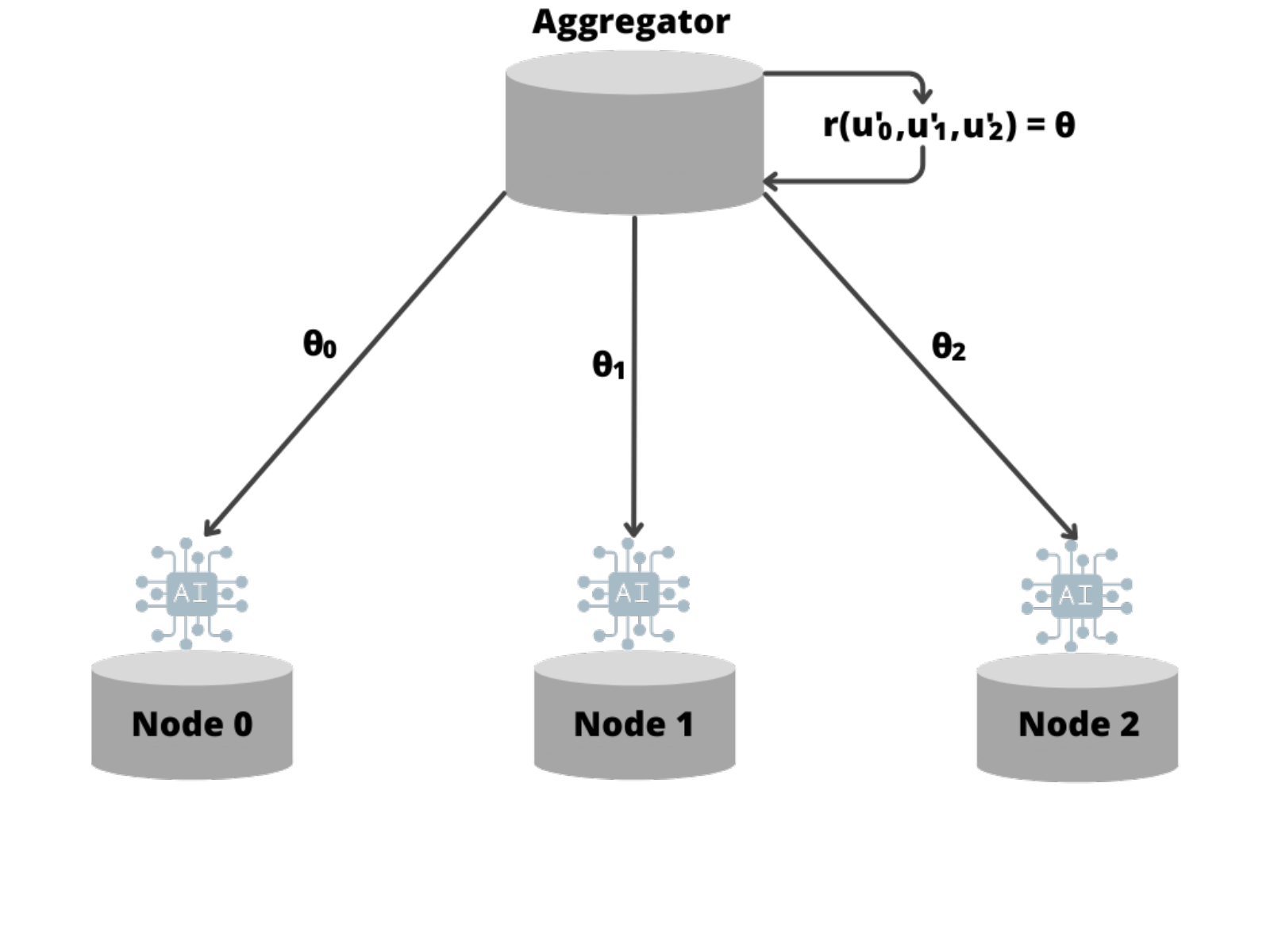}
    }
    \caption{PBSS phases}
    \label{fig:bss_phases}
\end{figure*}

Suppose a collection of $N$ nodes, each one having a private data vector
$\mathbf{X}^{(i)}=  \begin{pmatrix} X_{0}^{(i)} & \hdots & X_{K-1}^{(i)} \end{pmatrix}^T$,
where the input data $\mathbf{X}^{(i)} \in \mathbb{R}^{K \times L}$ is owned by node $i$, for $i \in \{0, \dots, N-1\}$. The PBSS consists of three phases, that are detailed as follows:

\underline{\textbf{Phase 1 - Sharing}:}
The first phase is based on classical Shamir Secret Sharing~\cite{ShamirSecretSharing:S79}. 
The client node $i$ composes the following encoded polynomial quotient
\begin{equation}
\begin{aligned}
    \label{eq:BSS}
    u_{\mathbf{X}^{(i)}}(z) &= \sum_{j = 0}^{K - 1} \frac{\frac{(-1)^j}{(z - \alpha_j)}}{\sum_{k = 0}^{K + T - 1} \frac{(-1)^k}{(z - \alpha_k)}} X_j^{(i)} + \sum_{j = 0}^{T-1} \frac{\frac{(-1)^{j + K}}{(z -\alpha_{j})}}{\sum_{k = 0}^{K + T - 1}\frac{(-1)^j}{(z - \alpha_k)}} R_j^{(i)},
\end{aligned}
\end{equation}
for some distinct interpolation points associated to the the data $\mathbf{X}^{(i)}$, 
$\boldsymbol\alpha = (\alpha_0, \dots, \alpha_{K-1}) \in \mathbb{R}^K$, which we choose 
again as the Chebyshev  points of first kind 
$\alpha_j = \cos\left(\frac{(2j + 1)\pi}{2K}\right)$, and some 
distinct interpolation points associated to the $R^{(i)}_j$, $\alpha_K, \dots, 
\alpha_{K + T - 1}  \in \mathbb{R}$, which are selected as the shifted Chebyshev points 
of first kind $\alpha_{K + j} = b + \cos\left(\frac{(2j + 1)\pi}{2K}\right)$. The 
rational function $u_{\mathbf{X}^{(i)}}(z)$ is evaluated at the set 
$\{z_j \}$, for $j = 0, \dots, N-1$, using $z_j = \cos(\frac{j\pi}{N - 1})$,
the Chebyshev points of second kind. Finally, the random coefficients $R_j^{(i)}$ are 
distinct  $1 \times L$ vectors drawn from a Gaussian distribution $\mathcal{N}(0, 
\frac{\sigma_n^2}{T})$. The evaluation $u_{\mathbf{X}^{(i)}}(z_j)$
is the share created from the client node $i$ and sent to the node $j$, as explained below.
The sharing phase is represented in Fig.~\ref{fig:bss_sharing_phase} assuming there are
three workers owning their own private data set ($X_j$) and locally computing an AI model.
As it is shown, the aggregator sends the global model ($\theta$) to the nodes, so they can
complete a new round of their learning task. The parameters of the global model are
codified, using the function $u()$ established in eq. \eqref{eq:BSS}, and shared with the
three nodes. After this codification part, the PBSS scheme acts: each node $j$ receives
the shares created using the evaluation point $z_j$, and computes the aggregation function
$f()$ over the codified parameters $u_{\theta'_i}(z_j)$. In this case, the input data to
be secured for the node $i$ is $\theta'_i$, the model updates after the local computation
is done. 

\underline{\textbf{Phase 2 - Computation and communication}:}
In this phase, the node $i$ calculates an arbitrary function $f$ using a set of 
polynomial quotient evaluations shared from the rest of nodes
$\{ u_{\mathbf{X}^{(0)}}(z_i), u_{\mathbf{X}^{(1)}}(z_i), \dots,  u_{\mathbf{X}^{(N-1)}}
(z_i)\}$,
where $u_{\mathbf{X}^{(j)}}(z_i)$ is the evaluation of the rational function 
corresponding to the input $X^{(j)}$ owned by the $j$-th node, and shared with 
node $i$.  The client node $i$ computes $f\bigl( u_{\mathbf{X}^{(0)}}(z_i),
u_{\mathbf{X}^{(1)}}(z_i), \dots, u_{\mathbf{X}^{(N-1)}}(z_i) \bigr)$
and sends the result to the master node. This phase is schematically represented 
in Fig.~\ref{fig:bss_computation_communication_phase}, where the obtained results
$f\bigl(u_{\theta'_0}(z_j), u_{\theta'_1}(z_j), u_{\theta'_2}(z_j)\bigr)$
are sent to the aggregator. 


\underline{\textbf{Phase 3 - Reconstruction}:}
In this last phase, the aggregator node (i.e., the master) reconstructs the value of the 
objective function over all the inputs using the results obtained from the subset 
$\mathcal{F}$ of the fastest workers, computing the reconstruction function
\begin{equation}
    \label{eq:reconstruction}
    \begin{aligned}
        &r_{\mathrm{Berrut}, \mathcal{F}}(z) = \sum_{i = 0}^{n} \frac{\frac{(-1)^i}{(z - 
        \tilde{z}_i)}}{\sum_{j = 0}^{n} \frac{(-1)^j}{(z - \tilde{z}_j)}} f\bigl( u_{\mathbf{X}^{(0)}}(\tilde{z}_i),
        u_{\mathbf{X}^{(1)}}(\tilde{z}_i), \dots, u_{\mathbf{X}^{(N-1)}}(\tilde{z}_i) \bigr),\\
    \end{aligned}
\end{equation}
where $\tilde{z}_i \in \mathcal{S} = \{ \cos\frac{j\pi}{N-1}, j \in \mathcal{F} \}$ 
are the evaluation points of the fastest nodes.
At this moment, the master node finds the approximation
$f\bigl(X^{(0)}_j, X^{(1)}_j, \dots, X^{(N-1)}_j\bigr) \approx r_{\mathrm{Berrut}, \mathcal{F}}(\alpha_j)$,
for all $j \in \{0, \dots, K - 1\}$.

Fig.~\ref{fig:bss_reconstruction_phase} shows the reconstruction phase for a secure aggregation.
When the aggregator receives the results $f(u_{\theta'_0}(z_j), u_{\theta'_1}(z_j),
u_{\theta'_{2}}(z_j))$ from each node, it can apply
the decoding function $r()$ to decode the global model. After that, the aggregator would compute
some arbitrary aggregation algorithm, e.g., SGD or any of its many variants in the literature
(FedAvg~\cite{McMahan2016}, 
SCAFFOLD~\cite{Karimireddy2020} or FedGen~\cite{Zhu2021}). Finally, the aggregator sends the updated
global model to the worker nodes, so the process can be repeated until convergence is achieved.

\subsection{Computational Complexity}
\label{sec:bss_scalability}


\subsubsection{Sharing} In this phase, a client node $i$ computes two sets of points $Z$ and 
$\hat{Z}$ as
\begin{align*}
    Z(j) &= \cos\left(\frac{(2j + 1)\pi}{2(K+T)}\right), \quad j = 0, \dots, K + T - 1, \\
    \hat{Z}(j) &= \cos\left(\frac{j\pi}{N - 1}\right), \quad j = 0, \dots, N - 1,
\end{align*}
so the complexity is $\mathcal{O}(K + T + N)$. Then, the node encodes the input data
via~\eqref{eq:BSS} for every value of $z = \{0, \dots, N-1\}$, so having a 
complexity of $\mathcal{O}(N(K + T))$. Finally, $N$ messages are sent from node 
$i$ to the rest of the nodes, which implies that the total number of messages 
$M$ exchanged in the network grows quadratically with the total number of nodes, i.e., 
$M = N (N - 1)$, with the size of the each message proportional to the number of 
columns $L$ of the input.

\subsubsection{Computation and Communication} In this phase, client node $i$ computes 
$f\bigl(u_{X^{(0)}}(z_i), u_{X^{(1)}}(z_i), \dots, u_{X^{(N-1)}}(z_i)\bigr)$, for the
arbitrary function $f$, with $u_{X^{(j)}}(z_i)$ a coded share of size $1 \times L$
received by node $j$, thus with a $\mathcal{O}(NL)$ complexity.
This is followed by transmission of the result to the master
node, so $N$ messages are exchanged at most, each one with size proportional to $L$.

\subsubsection{Reconstruction:} In this phase, the master node decodes the final 
result using a subset $\mathcal{F}$ of the results submitted by the clients, computing the 
function~\eqref{eq:reconstruction} over the set $\mathcal{F}$ of received responses,
which grows with $K$, $L$ and $n = |\mathcal{F}|$. This implies 
that the complexity of the decoding step is $\mathcal{O}(K(2n + L))$.

Since $\text{len}(Z) = K + T$, the total computational complexity is 
$\mathcal{O}(K(2n + L) + 3K + 3T + N + NL)$. For the communication complexity, 
$N (N - 1) + N$ messages of a size proportional to $L$ are exchanged at most.

\subsection{Comparison with HE and SMPC schemes}
\label{sec:pbss_comparison}

\begin{table}[t]
    \centering
    \caption{\label{table:pbss_comparison} Secure non-linear distributed computing schemes comparison.}
    \begin{tabular}{lccc} \\ 
    \textsc{protocol} & \textsc{privacy} & \textsc{precision} & \textsc{functions}  \\ \hline
    cons\_ltz~\cite{Aly22} & Information-theoretic & Exact & ReLU \\
    lss\_ltz~\cite{Aly22} & Information-theoretic & Exact & ReLU \\
    \cite{Zheng23} for Softmax & Information-theoretic & Exact & Softmax \\
    \cite{Zheng23} for Sigmoid & Information-theoretic & Exact & Sigmoid \\
    \cite{Lee23} & Computational & Approximate & ReLU \\ \hline
    PBSS & Bounded Leakage & Approximate & Any \\ \hline
    \end{tabular}
\end{table}

In \cite{Aly22}, the authors propose some protocols based on improvements of Rabbit~\cite{rabbit21} for
the computation of ReLUs, and test them for different configurations. Fastest protocols cons\_ltz and
lsss\_ltz require several rounds of communication, $R=4$ and $R=7$ respectively, with $N^2$ messages exchanged
for each of them. PBSS only requires $N^2$ messages at most with reduced size ($1 \times L$) for any non-linear function
computation.
Regarding computation complexity, PBSS only needs to compute the ReLU itself for reduced inputs of size $1 \times L$,
while these protocols have to evaluate a Garbled Circuit of approximately $G=200$ gates over normal size inputs,
which makes PBSS also outperform it. It is also worth mentioning that some of these protocols require circuit
optimizers to reduce the number of communication rounds, which impose a really high memory
usage to load the dependency graph. In \cite{Zheng23}, they propose a SMPC framework for the efficient
computation of Softmax and Sigmoid. For the case of Softmax, $2r$ rounds are required, being $r$ the number
of iterations (increasing $r$ results in greater accuracy, e.g., $r =\{8, 16, 32, 64\}$), with a complexity of
$(8mbr + 2b)rKL$, being $m$ the number of classes, and $b$ the bit-length of fixed point numbers. For the case
of Sigmoid, only $1$ round is required with a complexity of $12bKL$ at most. Similarly to what happens with Softmax,
PBSS outperforms them in communication complexity due to only requiring one round, and reducing the size of
the messages. Regarding computation, since we can directly compute the non-linear functions without making any
approximation for a polynomial, PBSS also outperforms them. In \cite{Lee23}, they propose a very
efficient approximation of the ReLU function for homomorphic evaluation that requires from $P \in [7, 8, \dots, 30]$
multiplications depending on the precision required. Since they are testing it with CKKS, the size of the
ciphertext scales linearly with the multiplicative depth, which makes PBSS outperforms it in computation times.
Regarding communication, this solution will always outperform PBSS, since it does not require to perform several
rounds of communication for evaluation or decryption. Additionally, it is important to highlight that we are comparing
our solution with schemes specifically dedicated and optimized for the execution of one or two types of non-linear functions,
while PBSS can compute any arbitrary function. The main downsides of PBSS against all solutions compared is that:
\begin{enumerate}
    \item It provides a weaker notion of privacy (bounded privacy leakage guarantee), contrary to the other solutions,
    that provide either information-theoretic privacy or computational privacy guarantees.
    \item It presents a high precision loss in small networks that require to share massive inputs. Regarding this
    issue, we have proposed two features to tackle it in~\ref{sec:matrix_mul_massive_inputs}.
\end{enumerate}

\section{Matrix multiplication with PBSS}
\label{sec:matrix-multiplication}

The other key primitive operation that our proposed framework should support for FL 
is the matrix multiplication.  Interpolation with rational functions, like in the Berrut 
approximation, does not allow  the product of matrices in a straightforward way due to the 
fact that encoding compresses the input matrix, making linear combinations of its rows. 
We show in this Section how  to adapt BACC to enable the approximate multiplication of 
two real matrices, both without and with privacy guarantees. The metric proposed to measure 
the privacy leakage  of  the matrix multiplications is also defined.

However, our method has two main limitations: (i) the error of the reconstruction is too
high when the number of worker nodes is lower than the number of rows of the input matrices
(this is inherited from the original BACC scheme), and (ii) high communication
costs in comparison to PBSS without privacy, from now on, BSS.
As massive input matrices are common in machine learning, these two limitations might hinder
the utility of the technique for large models. We mitigate the aforementioned problems using
two ideas:
first, by encoding multiple rows of a matrix in the same interpolation point; and secondly,
by using block matrix multiplication for large matrices, where the input matrix is
partitioned in vertical blocks.

\subsection{Approximate Matrix Multiplication without privacy}
\label{sec:matrix_mul_no_priv}

The proposed method has the following steps:

\subsubsection{} 
Each of the rows of the input matrices $A$ and $B$ are encoded using the rational function 
    $u_i(z) = \hat{q}_i(z) X_i$, being $X_i$ the row $i$ of the matrix to encode, and
    \begin{equation*}
      \hat{q}_i(z) = \frac{\frac{(-1)^i}{(z - \alpha_i)}}{\sum_{j = 0}^{K - 1}\frac{(-1)^j}{(z -
      \alpha_j)}}, \quad i = 0,1, \dots, K-1.
    \end{equation*}   
    \color{black}
    But instead of adding the rows of the matrices together, we keep them in place, 
    thus composing coded shares of the same size as the original matrix. Hence, a share 
    $u(\tilde{z})$ corresponding to the point $\tilde{z}$ would be composed as follows:
    $u(\tilde{z}) = 
        \begin{pmatrix}
            u_0(\tilde{z}) & u_1(\tilde{z}) & \hdots & u_{K-1}(\tilde{z})
        \end{pmatrix}^T$.

\subsubsection{}
Node $k$ receives two coded shares $u^{(A)}(z_k)$ and 
    $u^{(B)}(z_k)$ corresponding to the two input matrices
    $A = \begin{pmatrix} A_0 & A_1 & \hdots & A_{K-1} \end{pmatrix}^T$ and
    $B =\begin{pmatrix} B_0 & B_1 & \hdots & B_{K-1} \end{pmatrix}^T$ for
    $A_i \in \mathbb{R}^{1 \times d}$, $B_i 
    \in \mathbb{R}^{1 \times \ell }$. 
    Once the node $k$ has the operands, it computes the matrix multiplication so
    $f(u^{(A)}(z_k), u^{(B)}(z_k)) = u^{(A)}(z_k) \bigl( u^{(B)}(z_k) \bigr)^T$.

\subsubsection{}
Node $k$ divides each column $j$ of $u^{(A)}(z_k) \bigl( u^{(B)}(z_k) \bigr)^T$ 
    by the factor $\hat{q}_j(z_k)$, so
        \begin{equation} 
            \label{eq:matrix_mul}
        \begin{aligned}
            &g(u(z_k)) = \begin{pmatrix}
                    \frac{u_0^{(A)}(z_k) \bigl( u_0^{(B)}(z_k) \bigr)^T}{\hat{q}_0(z_k)} & \hdots 
                    & \frac{u_0^{(A)}(z_k) \bigl( u_{K-1}^{(B)}(z_k) \bigr)^T}{\hat{q}_{K-1}(z_k)} \\
                    \frac{u_1^{(A)}(z_k) \bigl( u_0^{(B)}(z_k) \bigr)^T}{\hat{q}_0(z_k)} & \hdots 
                    & \frac{u_1^{(A)}(z_k) \bigl( u_{K-1}^{(B)}(z_k) \bigr)^T}{\hat{q}_{K-1}(z_k)} \\
                    \vdots & \ddots & \vdots \\
                    \frac{u_{K-1}^{(A)}(z_k) \bigl( u_0^{(B)}(z_k) \bigr)^T}{\hat{q}_0(z_k)} & \hdots 
                    & \frac{u_{K-1}^{(A)}(z_k) \bigl( u_{K-1}^{(B)}(z_k) \bigr)^T}{\hat{q}_{K-1}(z_k)} \\
                \end{pmatrix}
        \end{aligned}
        \end{equation}
            for all $j = 0, \dots, K - 1$. Since $u_i^{(A)}(z_k) = \hat{q}_i(z_k) A_i$, 
            we can rewrite~\eqref{eq:matrix_mul} as
            \begin{equation*}
            \begin{aligned}
                &g(u(z_k)) = \begin{pmatrix}
                    \frac{\hat{q}_0(z_k) A_0 \hat{q}_0(z_k) B_0^T}{\hat{q}_0(z_k)} & \hdots & \frac{\hat{q}_0(z_k) 
A_0 \hat{q}_{K-1}(z_k) B_{K-1}^T}{\hat{q}_{K-1}(z_k)} \\
                    \frac{\hat{q}_1(z_k) A_1 \hat{q}_0(z_k) B_0^T}{\hat{q}_0(z_k)} & \hdots & \frac{\hat{q}_1(z_k) 
A_1 \hat{q}_{K-1}(z_k) B_{K-1}^T}{\hat{q}_{K-1}(z_k)} \\
                    \vdots & \ddots & \vdots \\
                    \frac{\hat{q}_{K-1}(z_j) A_{K-1} \hat{q}_0(z_k) B_0^T}{\hat{q}_0(z_k)} & \hdots & \frac{\hat{q}_{K-1}(z_k) A_{K-1} \hat{q}_{K-1}(z_k) B_{K-1}^T}{\hat{q}_{K-1}(z_k)}
                \end{pmatrix},
            \end{aligned}
            \end{equation*}
            which simplifies to
            \begin{equation*}
            \begin{aligned}
                & \begin{pmatrix}
                    \hat{q}_0(z_k) A_0 B_0^T & \hdots & \hat{q}_{0}(z_k) A_0 B_{K-1}^T \\
                    \hat{q}_1(z_k) A_1 B_0^T & \hdots & \hat{q}_{1}(z_k) A_1 B_{K-1}^T \\
                    \vdots & \ddots & \vdots \\
                    \hat{q}_{K-1}(z_k) A_{K-1} B_0^T & \hdots & \hat{q}_{K-1}(z_k) A_{K-1} B_{K-1}^T
                \end{pmatrix} =\operatorname{diag} \bigl( \hat{q}_0(z_k), \dots, \hat{q}_{K - 1}(z_k) \bigr)
                \begin{pmatrix}
                    A_0 B_0^T & \hdots & A_0 B_{K-1}^T \\
                    A_1 B_0^T & \hdots & A_1 B_{K-1}^T \\
                    \vdots & \ddots & \vdots \\
                    A_{K-1} B_0^T & \hdots & A_{K-1} B_{K-1}^T
                \end{pmatrix}
                \end{aligned}
            \end{equation*}

\subsubsection{}
Node $k$ sums all the rows of $g(u_{z_k})$ into one, reconstructing the original 
    compressed Berrut form
    \begin{equation}
        \label{eq:matrix_mul_updated}
        f(u(z_k)) =
            \begin{pmatrix}
                \sum_{i = 0}^{K - 1} \hat{q}_i(z_k) A_i B_0^T & \hdots & \sum_{i = 0}^{K - 1} \hat{q}_i(z_k) A_i B_{K-1}^T
            \end{pmatrix},
    \end{equation}
    and shares the result with the central node, which decodes the final result using the 
    same decoding function as the original scheme
    \begin{equation*}
        r_{\mathrm{Berrut}, \mathcal{F}}(z) = \sum_{i \in \mathcal{F}} 
        \frac{\frac{(-1)^i}{(z - \tilde{z}_i)}}{\sum_{j = 0}^{n}\frac{(-1)^j}{(z -
        \tilde{z}_j)}} f(u(\tilde{z}_i)).
    \end{equation*}

    Recalling the explanation of BACC in Section~\ref{sec:pbacc}, what
    the decoding operation does is to approximately calculate the value of the function at the
    points $\alpha_0, \alpha_1, \dots, \alpha_{K-1}$, so $r_{\mathrm{Berrut}, \mathcal{F}}(\alpha_i)
    \approx f(X_i)$. In order to achieve successful decoding work, it is required that 
    $u(\alpha_i) = X_i$. As the shares have the form $\sum_{i = 0}^{K - 1} \hat{q}_i(z_j) X_i$, 
    this is equivalent to saying that, when $u$ is evaluated in $\alpha_i$, $\hat{q}_i(z_j) = 1$, 
    and the rest of $\hat{q}_k(z_j) = 0$ for all $k \neq i$. For applying this to our matrix
    multiplication, we can easily check using~\eqref{eq:matrix_mul_updated} that if 
    $\hat{q}_i(z_j) = 1$ and $\hat{q}_k(z_j) = 0$ for all $k \neq i$, the resulting row is
    $f \bigl( u(z_j) \bigr)[\alpha_i] = \begin{pmatrix} 
    A_i B_0^T & A_i B_1^T & \hdots & A_i B_{K-1}^T \end{pmatrix}$,
    which is exactly the value of row $i$ of $A B^T$, and this holds for any
    $i$.

\subsection{Matrix multiplication with privacy}
\label{sec:matrix_mul_priv}

The process to add privacy to matrix multiplications is identical to the one explained
previously for BSS. Nevertheless, the encoding step has to be modified in order to 
introduce the random coefficients.  To accomplish this, we use the same rational 
function proposed initially, $u_i(z) = q_i(z) X_i + q_{K+i}(z) R_{K + i}$, i.e., using
\begin{equation*}
    q_i(z) = \frac{\frac{(-1)^i}{(z - \alpha_i)}}{\sum_{j = 0}^{K + T}\frac{(-1)^j}{(z - \alpha_j)}},
\end{equation*}  
except that the rows $k$ of the input matrix are not added together, but remain in the 
original position, and the randomness terms are added directly to each row $k$ 
multiplied by  the factor $q_k(z_j)$. Specifically,
\begin{equation}
    \label{eq:matrix_mul_privacy}
    \begin{aligned}
    u(z_j) &= 
        \begin{pmatrix}
            u_0(z_j) & u_1(z_j) & \hdots & u_{K-1}(z_j)
        \end{pmatrix}^T = \begin{pmatrix}
            q_0(z_j) X_0 + q_0(z_j) q_K(z_j) R_0 \\
            q_1(z_j) X_1 + q_1(z_j) q_{K + 1}(z_j) R_1 \\
            \vdots \\
            q_{K - 1}(z_j) X_{K-1} + q_{K-1}(z_j) q_{K + T - 1}(z_j) R_{T-1}
        \end{pmatrix}.
    \end{aligned}
\end{equation}
Combining~\eqref{eq:matrix_mul} and~\eqref{eq:matrix_mul_privacy} and simplifying terms, 
we get
\begin{equation*}
    \begin{aligned}
    &f(u(z_j)) = \begin{pmatrix}
            q_0(z_j) A_0 B_0^T + \tilde{q}_{0,0}(z_j) & \hdots & q_0(z_j) A_0 B_{K - 1}^T + \tilde{q}_{0,K-1}(z_j) \\
            \vdots & \ddots & \vdots \\
            q_{K - 1}(z_j) A_{K - 1} B_0^T + \tilde{q}_{K - 1,0}(z_j) & \hdots & q_{K - 1}(z_j) A_{K-1} B_{K - 1}^T + \tilde{q}_{K - 1, K - 1}(z_j)
        \end{pmatrix},
    \end{aligned}
\end{equation*}
for $\tilde{q}_{i,j}(z) = q_i(z) q_{K + j}(z) A_i R^{(B)}_j + q_i(z) a_{K + j}(z) B^T_j 
R^{(A)}_i + q_i(z) q_{K + i}(z) q_{K + j}(z)R^{(A)}_{i} R^{(B)}_j$. Adding all the 
rows together, we end up with a share like this
\begin{equation}
\label{eq:matrix_mul_privacy_updated}
\begin{aligned}
    &f(u(z_j)) =\begin{pmatrix}
            \sum\limits_{i = 0}^{K - 1} q_i(z_j) A_i B_0^T + \sum\limits_{i = 0}^{K - 1} 
            \tilde{q}_{i,0}(z_j) & \hdots & \sum\limits_{i = 0}^{K - 1} q_i(z_j) A_i B_{K - 1}^T + \sum\limits_{i = 0}^{K - 1}\tilde{q}_{i, K - 1}(z_j)
        \end{pmatrix}.
\end{aligned}
\end{equation}
Now, we can check using~\eqref{eq:matrix_mul_privacy_updated} that, if $q_i(z_j) = 1$ 
and $q_k(z_j) = 0$ for all $k \neq i$, the resulting row is $f(u(z_j)) = ( A_i B_0^T,  A_i B_1^T,  
\hdots, A_i B_{K-1})^T$
which is exactly the value of the row $i$ of $AB^T$, and is satisfied for any $i$. It is 
worth to mention that only one random row is being added per row of data ($T = K$) to 
facilitate the calculation, but the conditions hold for any number of random coefficients.
The remaining steps of the multiplication procedure are identical to those defined in the
previous Section.

\subsection{Privacy Analysis}
\label{sec:matrix_mul_privacy_analysis}

Since the method developed for multiplying matrices using Berrut coding requires keeping 
the size of the original matrix, and is rooted in the addition of random coefficients to 
every row individually, the privacy analysis will be done on individual rows. Therefore, 
the function to be analyzed is
\begin{equation}
    \label{eq:row-encoding}
    u_{i,v}(z) = X_i q_i(z) + \sum_{j = K + i}^{K + i + v - 1} R_{j} q_j(z),
\end{equation}
where $i = 0, \dots, K-1$, and $v = 0, 1, 2, \dots$ denote the number of 
random coefficients added per row. Consequently, the total number of added 
coefficients  are $T = K \times v$, and the scaling factor $q_i(z)$ is
\begin{equation*}
    q_i(z) = \frac{\frac{(-1)^i}{(z - \alpha_i)}}{\sum_{j = 0}^{K + T - 1}\frac{(-1)^j}{(z -\alpha_j)}}.
\end{equation*}

We use $\mathcal{C} = \{ j_1, \dots, j_{c} \}$ as the set of $c$ colluding nodes, and the
corresponding matrix of the coded elements of $X_i$ as
$L_{\mathcal{C},X_i} \triangleq
    \begin{pmatrix}
        q_{i}(z_{j_1}) & q_{i}(z_{j_2}) & \hdots & q_{i}(z_{j_{c}})
    \end{pmatrix}^T$,
and
\begin{equation*}
    \tilde{L}_{\mathcal{C}, X_i} \triangleq 
    \begin{pmatrix}
        q_{K + i}(z_{j_1}) q_i(z_{j_1}) & \hdots & q_{K+i+v-1}(z_{j_1}) q_i(z_{j_1}) \\
        q_{K + i}(z_{j_2}) q_i(z_{j_2}) & \hdots & q_{K+i+v-1}(z_{j_2}) q_i(z_{j_2}) \\
        \vdots & \ddots & \vdots \\
        q_{K + i}(z_{j_{c}}) q_i(z_{j_{c}}) & \hdots & q_{K+i+v-1}(z_{j_{c}}) q_i(z_{j_{c}})
    \end{pmatrix},
\end{equation*}
of size $c \times v$. The resultant mutual information upper bound is therefore
\begin{equation*}
    I_L(c, i)\leq \max_{\mathcal{C}: |\mathcal{C}| = c} \log_2 | I_{c} + \frac{s^2T}{\sigma^2_n} \tilde{\Sigma}_{\mathcal{C}}^{-1} \Sigma_{\mathcal{C}} |,
\end{equation*}
where $\tilde{\Sigma}_{\mathcal{C}} = \tilde{L}_{\mathcal{C}} 
\tilde{L}_{\mathcal{C}}^{\dagger}$ and $\Sigma_{\mathcal{C}} = L_{\mathcal{C}} 
L_{\mathcal{C}}^{\dagger}$.
Like the arguments in Section~\ref{sec:measuring_privacy_of_pbacc}, this result is 
an upper bound that measures the privacy leakage of the scheme. In this case, the privacy 
metric refers to an individual row of $i$ of the input matrix and $c$ colluding nodes, 
so we define our privacy metric $\imath_L$ as the averaged version of the mutual
information
\begin{equation*}
  \label{eq:privacy_metric2}
    \imath_L = \frac{\sum_{i = 0}^{K - 1} I_L(c,i)}{K} < \epsilon.
\end{equation*}

\subsection{Large Matrix Multiplication}
\label{sec:matrix_mul_massive_inputs}

Multiplication of large matrices arises frequently and plays a key role in many
applications of machine learning and artificial intelligence. In this Section, 
we introduce two simple methods for coded distributed private matrix multiplications 
(also valid for coded computing of other functions)  when 
the input matrices are large in comparison to the number of nodes collaborating in 
the computation. A first technique for drastically cutting the computational 
complexity consist of encoding multiple rows of the matrices using the same 
interpolation point. Such change does not prevent the  scheme to still compute a 
function, and it allows to get better precisions
with a considerably lower number of worker nodes compared to the length of the inputs.
A second alternative technique is to split the input matrices
vertically in blocks. This has the major advantage of reducing substantially the
communication and computation overheads, since, as we have explained, arbitrary functions
(including matrix multiplications) can be computed distributively over encoded blocks, and
then reconstructed completely in the decoding step.

\subsubsection{Encoding Multiple rows per point}
\label{sec:matrix_multiple_rpp}

Until now, the PBSS scheme takes
as a basis the rational function~\eqref{eq:BSS}. A consequence of the fact that this
function encodes one row of the matrix in each interpolation point is that the
communications costs decrease, since each encoded share always has length $1$. However,
this advantage can become a problem in cases where the number of worker nodes is much
smaller than the length of the inputs, as the final result will present a huge loss on
precision.

A simple way to solve this is to assign a fixed number of rows $r$ to each 
interpolation  point, using the modified encoding function
\begin{equation*}
    u'(z) = \sum_{i = 0}^{P - 1} \mathbf{X}'_i t_i(z) + \sum_{i = P}^{P + S - 1} \mathbf{R}'_{i - K} t_i(z),
\end{equation*}
where
\begin{equation*}
\mathbf{X}'_i = \begin{pmatrix}
    X_{ri}\\
    X_{ri+1} \\
    \vdots \\
    X_{ri + r - 1}
\end{pmatrix},
\mathbf{R}'_i = \begin{pmatrix}
    R_{ri}\\
    R_{ri+1} \\
    \vdots \\
    R_{ri + r - 1}
\end{pmatrix},
\end{equation*}
with $P = K / r$, $S = T / r$, and
\begin{equation*}
    t_i(z) = \frac{\frac{(-1)^i}{(z - \alpha_i)}}{\sum_{j = 0}^{P + S -1}
    \frac{(-1)^j}{(z -\alpha_j)}}.
\end{equation*}

Thanks to this approach, it is possible to guarantee that $P < N$ by adjusting the number 
of rows per point, $r$. Aside from this change, the rest of the steps in encoding and 
decoding remain the same. Achieving a similar gain for matrix multiplications is not so
direct, because the encoded matrix has exactly the same size as the input matrix. In 
substitution of the basis function for each coded row used initially for coded private
multiplication, i.e.~\eqref{eq:row-encoding}
we suggest to use the modified version
\begin{equation*}
    u'_{i}(z) = \mathbf{X}'_i t_i(z) + \sum_{j = P + i}^{P + i + V} \mathbf{R}'_{j - P} 
    t_j(z),
\end{equation*}
with $P = K / r$ and $V = T / r$. This new embedding implies that the encoded matrix is now
$U'_{z_j} = (u'_0(z_j), u'_1(z_j), \dots, u'_{P-1}(z_j))^T$, so it follows that
\begin{equation*}
    u'_i(z_j) = \begin{pmatrix}
        t_{ri}(z_j) X_{ri} + t_{ri}(z_j) t_{K + ri}(z_j) R_{ri} \\
        t_{ri}(z_j) X_{ri + 1} + t_{ri}(z_j) t_{K + ri}(z_j) R_{ri + 1} \\
        \vdots \\
        t_{ri}(z_j) X_{ri + q - 1} + t_{ri}(z_j) t_{K + ri}(z_j) R_{ri + r - 1}
        \end{pmatrix}.
\end{equation*}
After the matrix multiplication is performed over the coded shares, and the readjustment 
of the result columns (cf. Section~\ref{sec:matrix_mul_no_priv}) are done, all the 
$u'_i(z_j)$  are added together for $i \in \{0, 1, \dots, P - 1\}$, so the final share has 
$r$ rows instead of $1$. We should remark that this approach can be used either for 
approximation of arbitrary functions or for matrix multiplication, equally. 

\subsubsection{Block partition of matrices}
\label{sec:matrix_block_division}

The second method proposed in this paper to reduce communication and computation costs 
of multiplying large matrices ---other matrix operations are supported as well---, 
is based on splitting the original matrices vertically (i.e., by columns) in blocks 
or submatrices. Accordingly, let us write the input matrices in the form
$A = \begin{pmatrix}
            \hat{A}_0 & \hat{A}_1 & \hdots & \hat{A}_{b - 1}
        \end{pmatrix}$, $B = \begin{pmatrix}
            \hat{B}_0 & \hat{B}_1 & \hdots & \hat{B}_{b - 1}
        \end{pmatrix}$,
where $\hat{A}_x$ and $\hat{B}_y$ are the blocks $x$ and $y$ contained in $A$ and $B$, 
respectively, each block having $h$ columns. The objective is to (approximately) compute 
$f(A, B) = A^T B$, which is clearly
\begin{equation*}
    A^T B \triangleq 
    \begin{pmatrix}
        \hat{A}_0^T \hat{B}_0 & \hdots & \hat{A}_0^T \hat{B}_{b-1} \\
        \hat{A}_1^T \hat{B}_0 & \hdots & \hat{A}_1^T \hat{B}_{b-1} \\
        \vdots & \ddots & \vdots \\
        \hat{A}_{b-1}^T \hat{B}_0 & \hdots & \hat{A}_{b-1}^T \hat{B}_{b-1}
    \end{pmatrix}.
\end{equation*}
We assume that there are at least $b^2$ client nodes in the system, so that each node
receives from the master two shares
\begin{equation*}
    \begin{aligned}
        u^{(\hat{A}_x)}(z_j) = \begin{pmatrix} u_{hx}^{(\hat{A}_x)}(z_j) & u_{hx+1}^{(\hat{A}_x)}(z_j) & \dots &  u_{hx+h-1}^{(\hat{A}_x)}(z_j)\end{pmatrix},\\
        u^{(\hat{B}_y)}(z_j) = \begin{pmatrix} u_{hy}^{(\hat{B}_y)}(z_j) & u_{hy+1}^{(\hat{B}_y)}(z_j) & \dots &  u_{hy+h-1}^{(\hat{B}_y)}(z_j)\end{pmatrix},
    \end{aligned}
\end{equation*}
corresponding $\hat{A}_x$ and $\hat{B}_y$, for all 
possible combinations of $x$ and $y$. Identifying node $j$ with its pair of indices 
$(x, y)$ (a specific block), it will calculate the share
\begin{equation*}
\begin{aligned}
    &\hat{f}_{(x,y)}\bigl(u^{(\hat{A}_x)}(z_j), u^{(\hat{B}_y)}(z_j)\bigr) =\begin{pmatrix}
        \frac{u_{hx}^{(\hat{A}_x)}(z_j) u_{hy}^{(\hat{B}_y)}(z_j)}{q_{hy}(z_j)} & \hdots & 
        \frac{u_{hx}^{(\hat{A}_x)}(z_j) u_{hy + h - 1}^{(\hat{B}_y)}(z_j)}{q_{hy + h - 1}
        (z_j)} \\
        \frac{u_{hx + 1}^{(\hat{A}_x)}(z_j) u_{hy}^{(\hat{B}_y)}(z_j)}{q_{hy}(z_j)} & \hdots 
        & \frac{u_{hx + 1}^{(\hat{A}_x)}(z_j) u_{hy + h - 1}^{(\hat{B}_y)}(z_j)}{q_{hy + h - 
        1}(z_j)} \\
        \vdots & \ddots & \vdots \\
        \frac{u_{hx + h - 1}^{(\hat{A}_x)}(z_j) u_{hy}^{(\hat{B}_{y})}(z_j)}{q_{hy}(z_j)} & 
        \hdots & \frac{u_{hx + h - 1}^{(\hat{A}_{x})}(z_j) u_{hy + h - 1}^{(\hat{B}_y)}(z_j)}
        {q_{hy + h - 1}(z_j)}
    \end{pmatrix},
\end{aligned}
\end{equation*}
and then adds together all the rows of $\hat{f}_{(x,y)}\bigl(u^{(\hat{A}_x)}(z_j), u^{(\hat{B}_y)}(z_j)\bigr)$ into a single value. 
The master node will use all the results of the block $(x, y)$ to decode 
$\hat{A}_x^T \hat{B}_y$. Once all the blocks are decoded, the matrix 
product can be reconstructed. Obviously, both the multiple encoding per point 
and block matrix multiplication as described in this Section can be jointly used.

\subsection{Complexity analysis}
\label{sec:matrix_mul_scalability}

The global computation and communication cost for coded private computation of matrices 
can be obtained as follows.

\subsubsection{Sharing phase} the cost is similar to normal PBSS 
(cf. Section~\ref{sec:bss_scalability}), but with only two inputs and the size of
the messages grows linearly 
with $K L$,  where $K$ and $L$ denote the rows and columns of the input matrix,
respectively. Two variants can be distinguished: (i) Encoding multiple rows in each 
interpolation point, for which the analysis is identical and the computational cost is unchanged;
(ii) Block partitioning: in this case the length of messages transmitted in the sharing phase 
is reduced, since the size of blocks is linear in $K H$, with  $H = L / b$ the number of columns 
of each block, and in $b$, the total number of vertical blocks in which the input matrix was divided.
Note that $H < l$ for $b > 1$.

\subsubsection{Computation and Communication phases} here, each node $i$ computes a matrix
product over two coded shares, with a computational complexity 
$\mathcal{O}(K L)$. After completing this step, the columns of the resulting 
share are re-scaled by a constant factor, which requires $\mathcal{O}(C)$ elementary 
operations assuming that $C$ is the number of columns, and all the rows of the 
resulting share are lumped together $\sum_{i = 0}^{K - 1 } u_i(z_w)$, which 
requires $K - 1$ additions. Hence, the total complexity is $\mathcal{O}(K + C)$. 
Note that, for calculating a single function, $C = L$, but for matrix 
multiplications $C = K$. Finally, each node shares the result with the master 
node, so $N$ messages are exchanged at most, of size proportional to $L$. 
We consider again the two possible enhancements of efficiency: (i)
Encoding multiple rows in each interpolation point. Not all rows are added together, 
so the sum gets reduced to complexity $\mathcal{O}(K/r)$, 
where $r$ is the number of rows encoded in each interpolation point. 
However, the size of the messages is now greater, equal to $r L$; (ii)
Block partitioning. The computation complexity is reduced, the blocks 
have $h$ columns instead of $L$, so the complexity is $\mathcal{O}(K h)$. 
The size of the messages is reduced as well, as the results shared among 
the nodes scale with $h = L / b$.

\subsubsection{Reconstruction phase} Similar to normal PBSS.

When multiple rows encode each interpolation point, 
the complexity of the decoding step now scales with $K/r$, $2n$ and $r L$, 
so the complexity is $\mathcal{O}(\frac{K}{r}(2n + r L))$.
Instead, dividing the inputs in vertical blocks, the complexity 
of the decoding step now scales with $b$, $K$, $2n$ and $h$, and it 
is $\mathcal{O}(b K(2n + h))$.
Therefore, the total computational complexity is 
$\mathcal{O}(K(2n + L) + K L + 3K + 3T + N + C)$. For the communication cost, 
assuming two input matrices, $2N$ messages are required, each one of a size that scales with $K L$, and $N$ 
messages with a size proportional to $L$, are exchanged at most. Analyzing the 
two features for dealing with large inputs, we obtain that: (i) with multiple rows 
per interpolation point: The total computational complexity is 
$\mathcal{O}(\frac{K}{r}(2n + r L)  + K L + 2K + K/r + 3T + N + C)$. 
Regarding communication, $2N$ messages with a size that scales with $K L$, 
and $N$ messages with a size that scales with $r L$, are exchanged at most; (ii) 
Input divided in vertical blocks: The total computational complexity is 
$\mathcal{O}(bK(2n + h) + Kh + 3K + 3T + N + C)$. As for communication, $2N$ messages with size $Kh$, and $N$ messages with size $h$, 
are exchanged at most.

\subsection{Comparison with other Secure Matrix Multiplication Schemes}
\begin{table}[t]
    \centering
    \caption{\label{table:} Secure distributed matrix multiplication schemes comparison.}
    \begin{tabular}{lcccc} \\ 
    \textsc{protocol} & \textsc{privacy} & \textsc{precision} & \textsc{partition} \\ \hline
    SCMPC~\cite{PolynomialSharing:NA21} & Information-theoretic & Exact & Vertical \\
    EPC~\cite{EntangledPolynomial:QA20} & Information-theoretic & Exact & Complex \\
    FPSCMM~\cite{KIM2023722} & Information-theoretic & Exact & Complex \\
    \hline
    PBSS & Bounded Leakage & Bounded Error & Vertical \\ \hline     
    \end{tabular}
\end{table}

In~\cite{PolynomialSharing:NA21}, they propose a scheme for secure massive matrix operations that provides a similar
communication complexity as PBSS in the sharing phase, since they both allow to partition
the input matrices in vertical blocks, but a worse result sharing phase, as PBSS compresses the result blocks to $1$
row before sending them. The main strengths of~\cite{PolynomialSharing:NA21} against PBSS is that provides
information-theoretical privacy in presence of colluding nodes, and an exact computation of the matrix result. In \cite{EntangledPolynomial:QA20} and \cite{KIM2023722}, they propose more efficient methods of secure matrix multiplication,
with the possibility of performing complex partitions, rather than just vertical ones. All of the schemes compared resist
stragglers better than PBSS as long as the number of colluding nodes or stragglers in the network is not too high, which
could impede the results to be reconstructed. In those cases, PBSS could have an advantage if the precision requirements are
not too high, as it always reconstructs the result with a given error. Another aspect that should be highlighted is that
all of the compared schemes rely on polynomial interpolation for decoding results, which leads numerical instabilities
when using real numbers or when the number of nodes of the network is too high.
However, the main strength that PBSS has against all the compared schemes is that it is able to compute both matrix
multiplications and non-linear functions (really common operations in machine learning). Having achieved a scheme that
can compute both operations means that we can deal with the decentralized training of machine learning models with the
same scheme. This means that we only need to do one sharing phase, one computation phase, and one decoding phase, and
we could reconstruct combinations of non-linear functions and matrix multiplications, while keeping the privacy of the
input data. To the extent of our knowledge, our scheme is the most efficient approach that can deal with the two
capabilities combined, which means that if we want to securely distribute the training of some ML model in a more efficient way
than using PBSS, we would need at least two different schemes (one for dealing with the secure computation of non-linear
functions, and the other for the secure computation of matrix multiplications). This imposes two main issues: (i) the
more schemes we have to deal with, the harder it is to integrate them in a ML model, and (ii) the schemes in this field
always require, at least, some sharing, computation and decoding phases, that would need to be repeated for both schemes
and for each function computed, increasing the communication and computational costs of the whole 
distributed task.

\color{black}

\section{Results}
\label{sec:results}

We provide in this Section numerical results obtained after performing two sets of experiments. 
First, computations of non-linear functions relevant to FL, comprising the developments of 
PBACC (Sect.~\ref{sec:pbacc}) and PBSS (Sect.~\ref{sec:berrut-secret-sharing}).
Specifically, we tested several common activation functions (ReLU, Sigmoid and Swish) and
also typical functions used in non-linear aggregation methods (Binary Step, and Median). 
Secondly, we conducted tests related to matrix multiplications, demonstrating the adaptations 
of PBSS for this setting (Sect.~\ref{sec:matrix-multiplication}). Matrix
products with normal and sparse matrices have been computed.

The experimental design is as follows. First, we fix a maximum privacy leakage $\epsilon$ in 
order to obtain the values of the other parameters to ensure that theprivacy level is
achieved in presence of a semi-honest minority of
colluding nodes ($c=50$, which represents $25$\% of total). These values are summarized in
Table~\ref{tab:experimentation_parameters}. After that, we measure the numerical precision in 
presence of stragglers with the aim of comparing this precision to the one obtained with the original 
BSS scheme (without any form of privacy). Our goal is to calculate the cost
(in terms of precision) of adding privacy. Finally, for non-linear computations we also compare BSS
with basic differential privacy: an alternative technique to compute non-linear functions. It is worth
mentioning that, since we work in a multi-input configuration, the actual computed result
$\hat{r}$ will be the addition of the function values over each
individual input, so $\hat{r} = \sum_{i = 0}^{N - 1}f(X^{(i)})$.



\begin{table}[t]
    \centering
    \caption{\label{tab:experimentation_parameters} Parameters for the experimental settings.}
    \begin{tabular}{clc}\\
    \textbf{Symbol} & \textbf{Description} & \textbf{Value} \\
    \midrule
    $N$ & Number of nodes & $200$ \\
    $K$ & Length of each input & $1000$ \\
    $s$ & Max value that any element of the input can take & $100$ \\
    $\sigma_n$ & Standard deviation of the randomness in PBSS & $10000$ \\
    $T$ & Number of random coefficients & $1000$ \\
    $c$ & Number of colluding nodes & $50$ \\
    $\sigma_{dp}$ & Standard deviation of randomness added in DP & $30$ \\
    $r$ & Rows per interpolation point & $50$ \\
    \bottomrule
    \end{tabular}
\end{table}


Working under a very pessimistic scenario, where all the encoded private information would be 
inferred if all the workers collude, the theoretical maximum privacy leakage  
is $\imath_L = H(X) = \log(2s) \approx 14.28$ bits. With the parameters listed in 
Table~\ref{tab:experimentation_parameters}, if a minority of  $50$ nodes collude, the maximum 
information that could be obtained is approximately $0.197$ bits, which means that, at least, 
$98.62\%$ of information remains secure. For FL settings this is a good security level, but 
the parameters in Table~\ref{tab:experimentation_parameters} can be adjusted accordingly if a 
tighter privacy is desired.



\subsection{Computation with non-linear functions}
\label{sec:results_nonlinear}

\begin{figure*}[t]
    \centering
    \subfloat[ReLU computation comparison.\label{fig:relu_computation_comparison}]{%
        \includegraphics[width=0.50\linewidth]{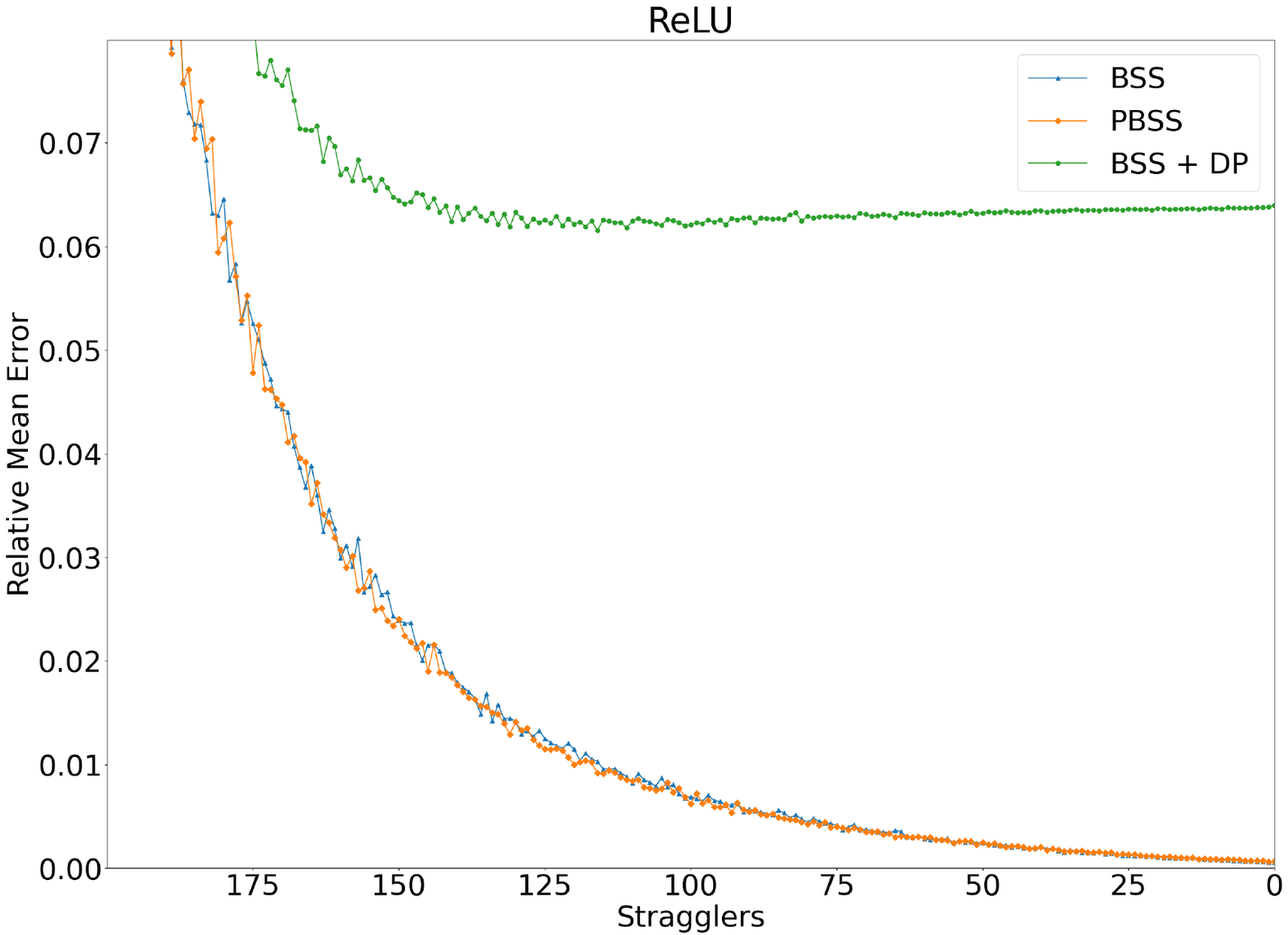}
    }
    \subfloat[Sigmoid computation comparison.\label{fig:sigmoid_computation_comparison}]{%
        \includegraphics[width=0.50\linewidth]{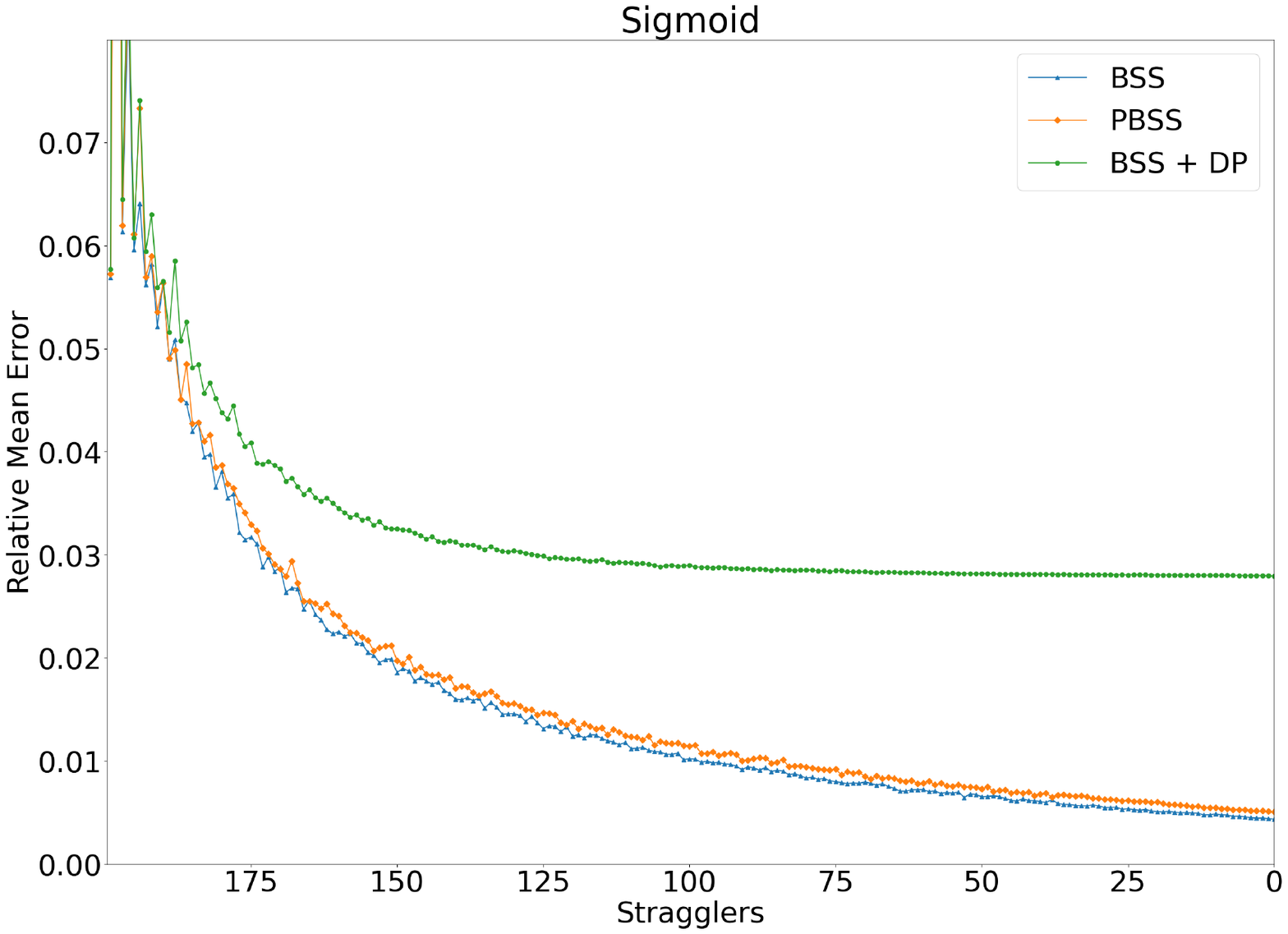}
    }
    \\
    \begin{tabular}{cccc}
    \toprule
    \textbf{Stragglers} & BSS & PBSS & BSS + DP \\
    \midrule
    $100$ & $0.006874037$ & $0.006209476$ & $0.062076736$ \\
    $50$ & $0.002417026$ & $0.002503581$ & $0.063198365$ \\
    $0$ & $0.000577592$ & $0.000655003$ & $0.063969884$ \\
    \bottomrule
    \label{tab:relu_computation_comparison}
    \end{tabular}
    \begin{tabular}{cccc}
    \toprule
    \textbf{Stragglers} & BSS & PBSS & BSS + DP \\
    \midrule
    $100$ & $0.010227901$ & $0.011458554$ & $0.028995183$ \\
    $50$ & $0.006560695$ & $0.007300909$ & $0.028202122$ \\
    $0$ & $0.004400018$ & $0.005110319$ & $0.027956606$ \\
    \bottomrule
    \label{tab:sigmoid_computation_comparison}
    \end{tabular}
    \caption{Numerical precision with activation functions ReLU and Sigmoid.}
    \label{fig:relu_sigmoid_swish_computation_comparison}
\end{figure*}

\begin{figure}
    \centering
    \subfloat[Swish computation comparison.\label{fig:swish_computation_comparison}]{%
        \includegraphics[width=0.5\linewidth]{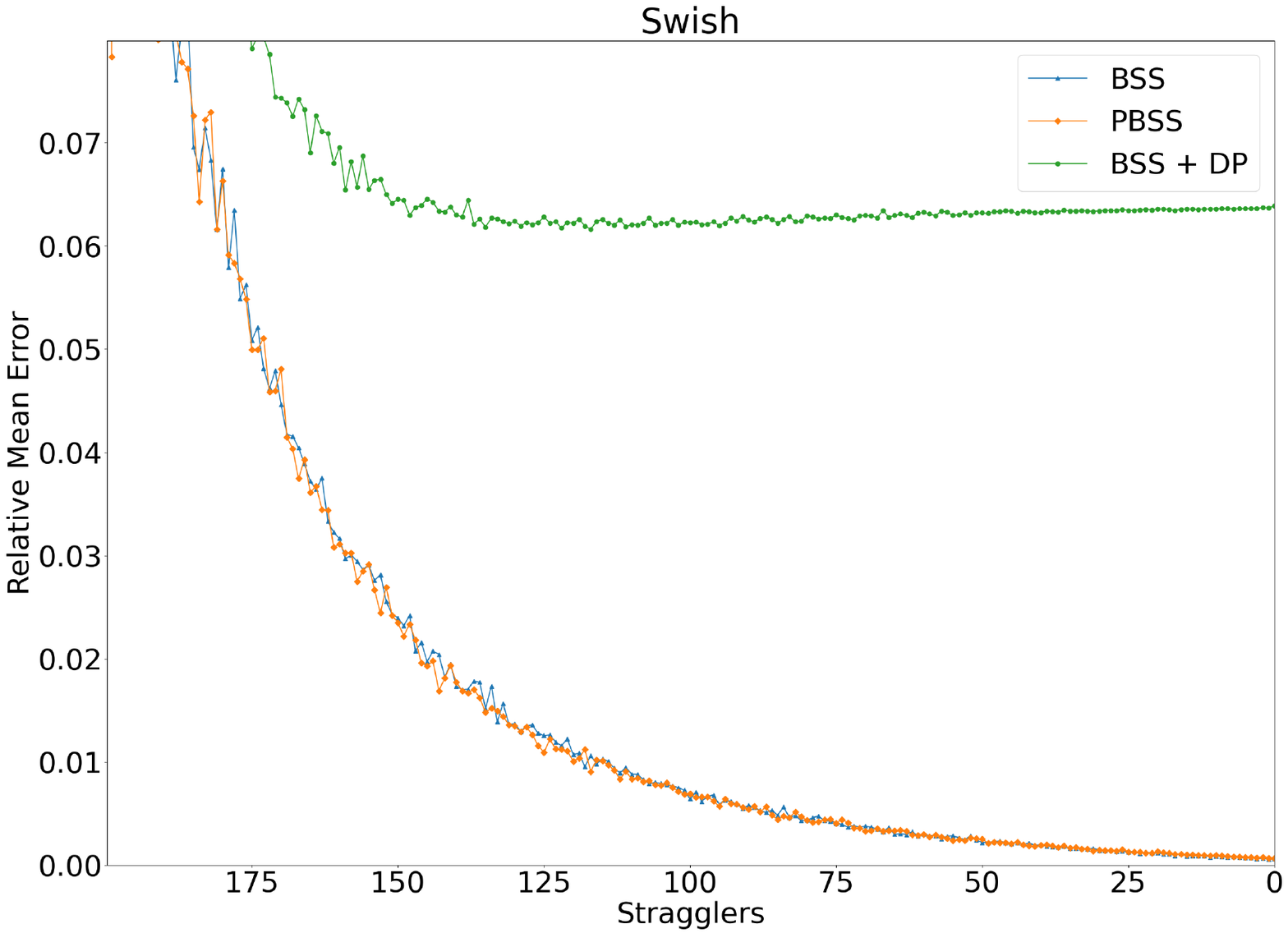}
    }
    \\
    \begin{tabular}{cccc}
    \toprule
    \textbf{Stragglers} & BSS & PBSS & BSS + DP \\
    \midrule
    $100$ & $0.006389195$ & $0.006893500$ & $0.062273931$ \\
    $50$ & $0.002165249$ & $0.002500792$ & $0.063192200$ \\
    $0$ & $0.000596237$ & $0.000675981$ & $0.063865781$ \\
    \bottomrule
    \label{tab:swish_computation_comparison}
    \end{tabular}
    \caption{Numerical precision with Swish activation function.}
     \label{fig:relu_sigmoid_swish_computation_comparison-bis}
\end{figure}

We have compared our proposal (PBSS) to PBSS without privacy (BSS) and BSS adding a differential privacy scheme (BSS + DP). 
As it was previously mentioned, we have tested our proposal with three non-linear functions that are often used as activation functions in machine learning solutions: ReLU (Fig.~\ref{fig:relu_computation_comparison}), Sigmoid (Fig.~\ref{fig:sigmoid_computation_comparison}) and Swish (Fig.~\ref{fig:swish_computation_comparison}). We have used the normalized relative mean error as performance metric
\begin{equation*}
    \text{RME} = \frac{ \| \frac{\tilde{Y} - Y }{Y} \|_1}{\text{len}(Y)},
\end{equation*}
\noindent where $Y$ is he exact result, $\tilde{Y}$ is its approximation, $\| \cdot \|_1$ is the entrywise $1$-norm of a matrix and $\text{len}()$ the total number of elements of a matrix (the division is done element by element).


Tables in Fig.\ref{fig:relu_sigmoid_swish_computation_comparison} show a summary of the the RME value when receiving different numbers of results from workers: $100$, $150$ and $200$, i.e. when the stragglers are $100$ nodes, $50$ nodes or $0$ nodes, respectively (note that the number of workers $N = 200$ is defined in Table~\ref{tab:experimentation_parameters}).

As expected, since ReLU and Swish are similar functions, their behaviours are also similar. In both cases PBSS and BSS improve the numerical precision when the number of received results increases. However, the results obtained when applying BSS plus differential privacy are not so good. They do not improve when the number of results is higher than $50$, offering less privacy ($\sigma = 30$) with no guarantees against colluding nodes. Therefore, the BSS scheme  (without privacy) offers slightly better results than PBSS, being this a little cost to pay to include privacy in the computation. Precisely, the cost can be measured in terms of percentage of accuracy loss, being approximately $0.008\%$ for ReLU, $0.071\%$ for Sigmoid, and $0.008\%$ for Swish.

According to the results shown in Fig.\ref{fig:relu_sigmoid_swish_computation_comparison}, the 
Sigmoid function seems to be harder to approximate than ReLU and Swish, since the precision obtained
is lower than in the other two functions under BSS and PBSS. The precision performs better with 
Sigmoid than with ReLU or Swish using BSS + DP when the number of stragglers is less than $50$, 
although it does not outperform the results of our proposal. 

\begin{figure*}[t]
    \centering
    \subfloat[Binary Step computation comparison.\label{fig:binary_step_computation_comparison}]{%
        \includegraphics[width=0.5\linewidth]{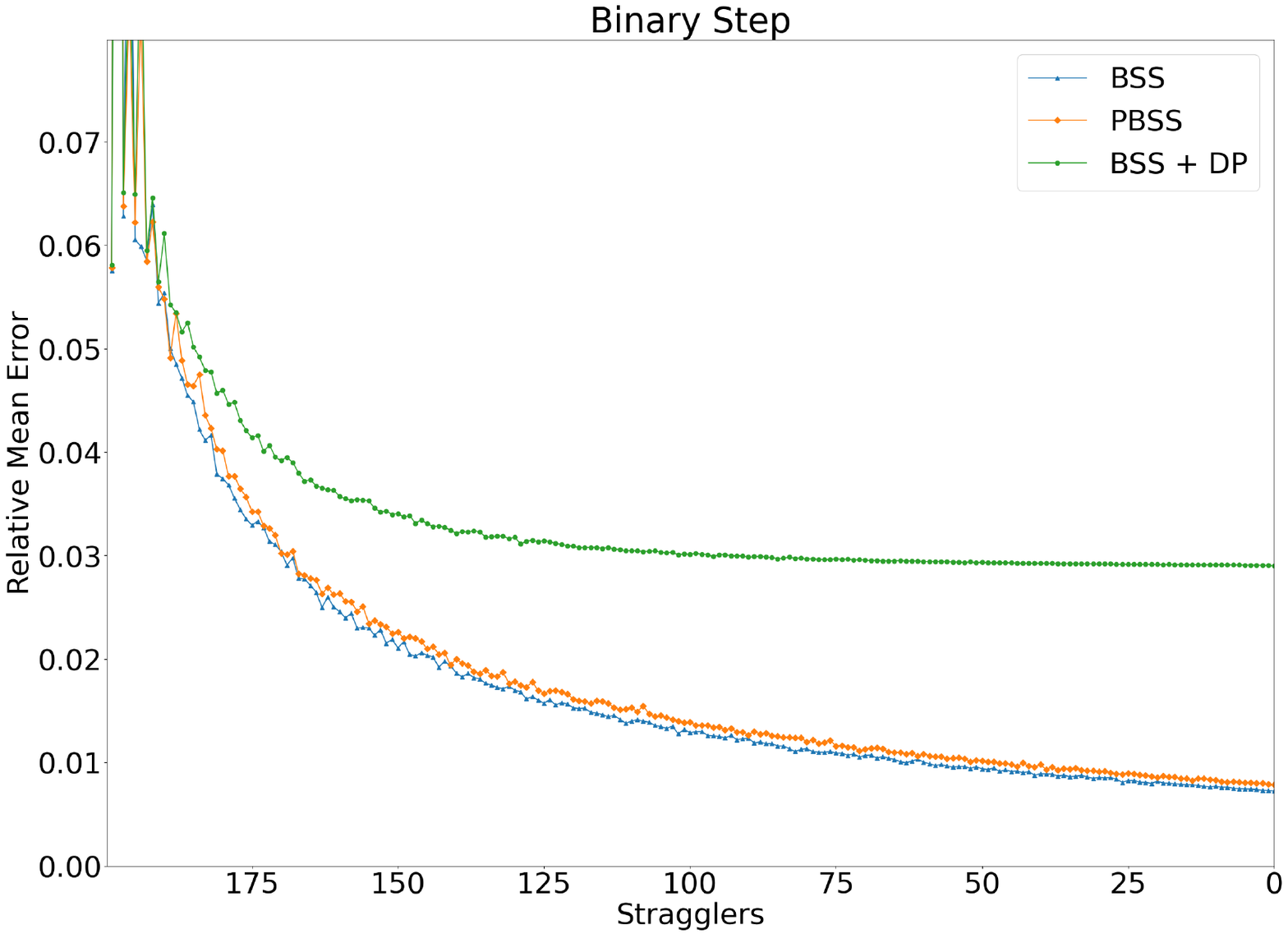}
    }
    \subfloat[Median computation comparison.\label{fig:median_computation_comparison}]{%
        \includegraphics[width=0.5\linewidth]{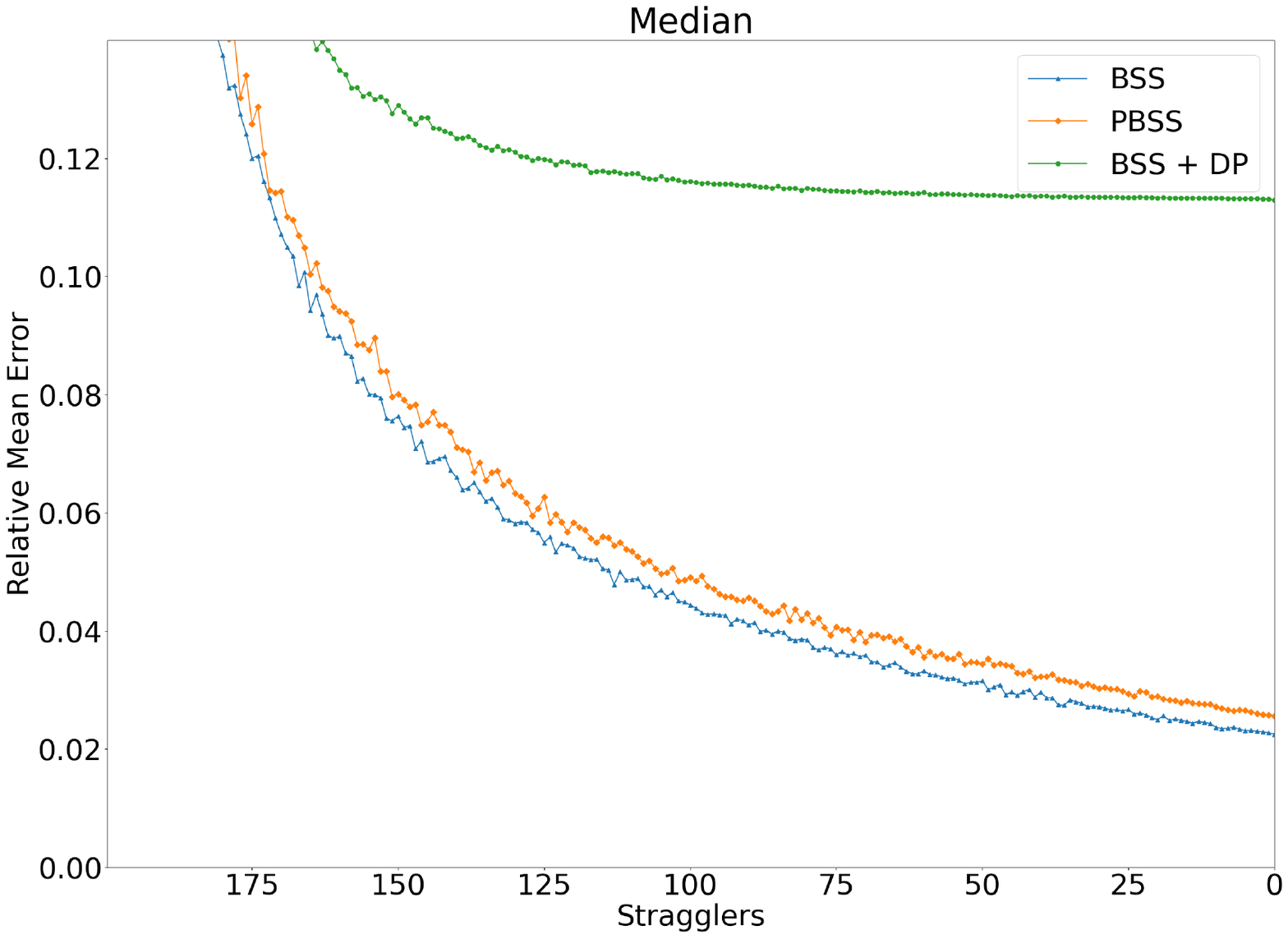}
    }
    \\
    \begin{tabular}{cccc}
    \toprule
    \textbf{Stragglers} & BSS & PBSS & BSS + DP \\
    \midrule
    $100$ & $0.012875861$ & $0.013881484$ & $0.030098946$ \\
    $50$ & $0.009368434$ & $0.010180803$ & $0.029364728$ \\
    $0$ & $0.007248344$ & $0.007854828$ & $0.029014532$ \\
    \bottomrule
    \label{tab:binary_step_computation_comparison}
    \end{tabular}
    \begin{tabular}{cccc}
    \toprule
    \textbf{Stragglers} & BSS & PBSS & BSS + DP \\
    \midrule
    $100$ & $0.044372558$ & $0.049100067$ & $0.116038774$ \\
    $50$ & $0.031538214$ & $0.034423612$ & $0.113759435$ \\
    $0$ & $0.022517222$ & $0.025564940$ & $0.112916117$ \\
    \bottomrule
    \label{tab:median_computation_comparison}
    \end{tabular}
    \caption{Precision results using different aggregation functions (Binary Step and Median). The Tables contain the relative errors.}
\end{figure*}

We have also tested the precision results when using two typical aggregation functions in FL: 
Binary Step and the Median. As Fig.~\ref{fig:median_computation_comparison} shows, Median is harder
to approximate than Binary Step (Fig.~\ref{fig:binary_step_computation_comparison}). Since the errors 
for the aggregation functions are larger than in the previous case, more rounds would be needed for 
reaching model convergence that without including privacy.



To sum up, the errors we have obtained for the activation functions with PBSS are quite close to 
the ones obtained without privacy (BSS), working with large matrices ($K=1000$) and for a 
reasonable privacy level ($\imath_L=0.197$ bits). Thus, PBSS could be suitable for secure 
federated learning for classic machine learning models (linear/logistic regressions, neural 
networks, convolutional neural networks, etc.). Besides, the results obtained for the two aggregation 
methods, suggest that PBSS would fit in FL settings where the objective is to securely aggregate 
a model resilient to malicious adversaries. Additionally, PBSS should also be able to approximate 
complex aggregations that require comparisons (Binary step), e.g., in survival analysis with Cox
regression. That said, as the errors for the aggregation functions are larger than in the previous 
case, it is likely that this will have an impact on the number of rounds it takes the model to converge.


\subsection{Testing Matrix Multiplications}
\label{sec:results_matrix}

We also tested our proposal for matrix multiplications with two types of operations: common 
matrix multiplications (two inputs) and matrix multiplication in blocks (two inputs divided 
in blocks). We computed the numerical precision in presence of stragglers to compare our PBSS 
with the original BSS (without privacy) in order to evaluate the trade-off between privacy and 
accuracy. The matrices were randomly generated with entries drawn from a uniform distribution.

\begin{figure*}[t]
    \centering
    \subfloat[Conventional Matrix multiplication.\label{fig:matrix_mul_computation_comparison}]{%
        \includegraphics[width=0.50\linewidth]{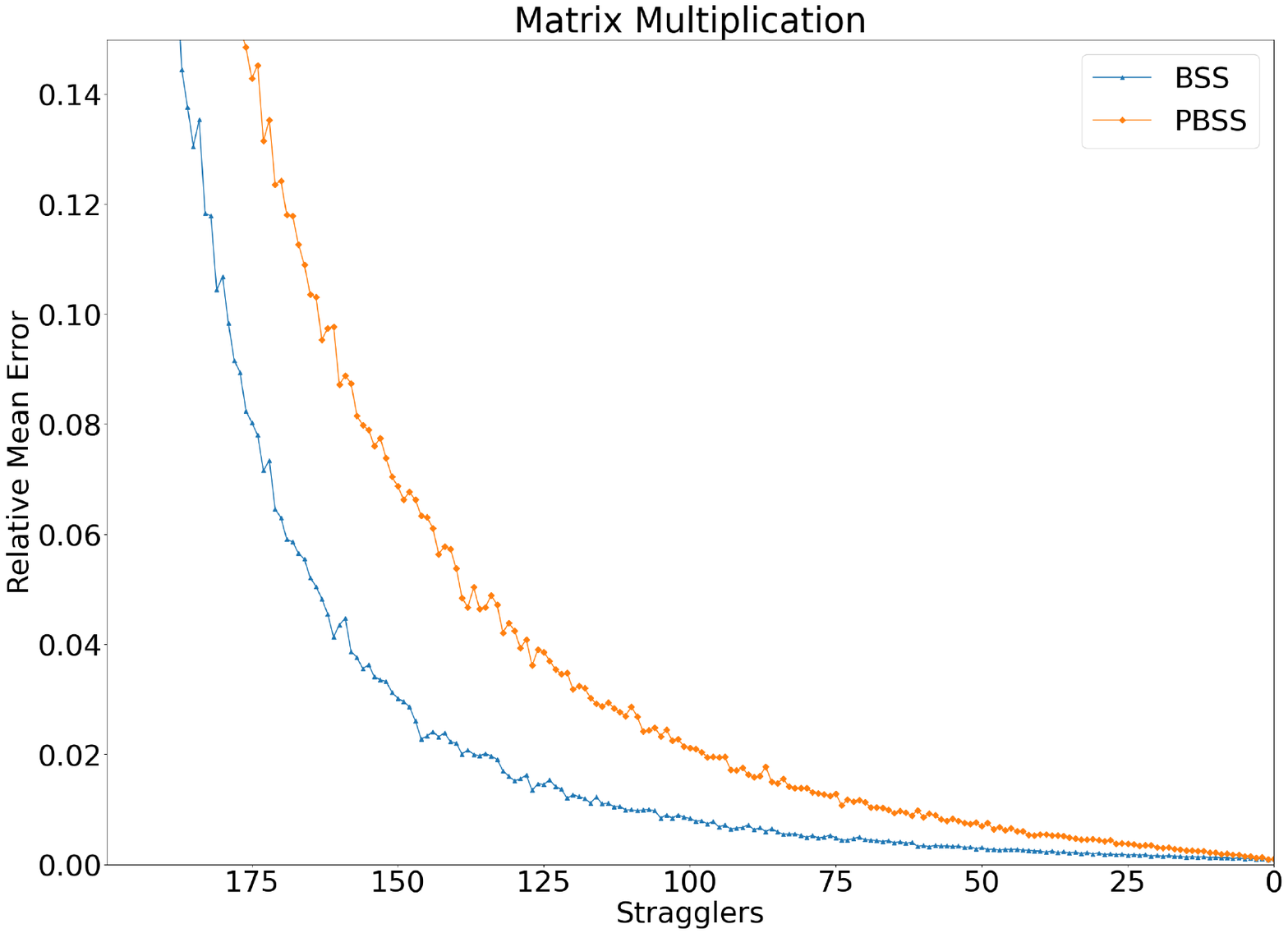}
    }
    \subfloat[Sparse Matrix Multiplication.\label{fig:sparse_matrix_mul_computation_comparison}]{%
        \includegraphics[width=0.50\linewidth]{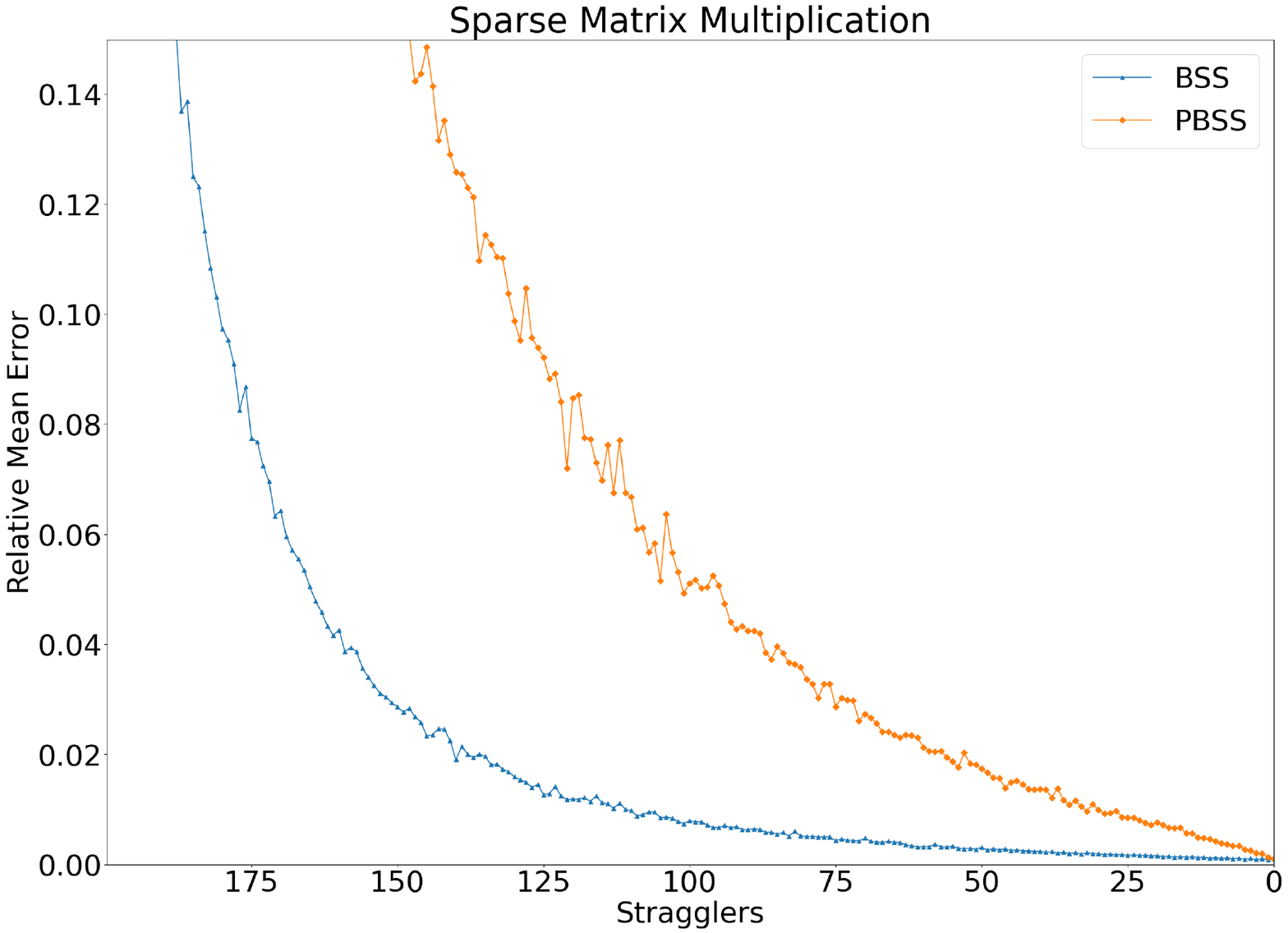}
    }
    \\
    \begin{tabular}{ccc}
    \toprule
    \textbf{Stragglers} & BSS & PBSS\\
    \midrule
    $100$ & $0.005357120$ & $0.031209873$ \\
    $50$ & $0.002866487$ & $0.012143143$ \\
    $0$ & $0.000995137$ & $0.001074013$ \\
    \bottomrule
    \label{tab:matrix_mul_computation_comparison}
    \end{tabular}
    \begin{tabular}{ccc}
    \toprule
    \textbf{Stragglers} & BSS & PBSS \\
    \midrule
    $100$ & $0.007925727$ & $0.051045426$ \\
    $50$ & $0.003098950$ & $0.017392449$ \\
    $0$ & $0.000970545$ & $0.001014285$ \\
    \bottomrule
    \label{tab:sparse_matrix_mul_computation_comparison}
    \end{tabular}
    \caption{Common Matrix Multiplications computation. The Tables show the relative errors.}
    \label{tab:mul_computation_comparison}
\end{figure*}

Fig.~\ref{tab:mul_computation_comparison} shows the precision results when using the common matrix multiplication scheme proposed in Section~\ref{sec:matrix_mul_priv} using conventional matrices (Fig.~\ref{fig:matrix_mul_computation_comparison}) and sparse matrices (Fig.~\ref{fig:sparse_matrix_mul_computation_comparison}. In both cases, PBSS offers almost identical results than the original scheme without privacy (BSS), so adding privacy does not have a high cost in performance. Anyway, it is worthy to mention that, as expected, the lower the number of stragglers the better the result achieved. But, differently to the results of the previous analysis (non-linear functions and aggregation functions in Section~\ref{sec:results_nonlinear}, BSS and PBSS perform worse in presence of stragglers, so they need more results to reduce the error. This behaviour is even more accentuated with sparse matrices. 

\begin{figure*}[t]
    \centering
    \subfloat[Conventional Matrix multiplication divided in 2 vertical blocks.\label{fig:matrix_mul_blocks_computation_comparison}]{%
        \includegraphics[width=0.50\linewidth]{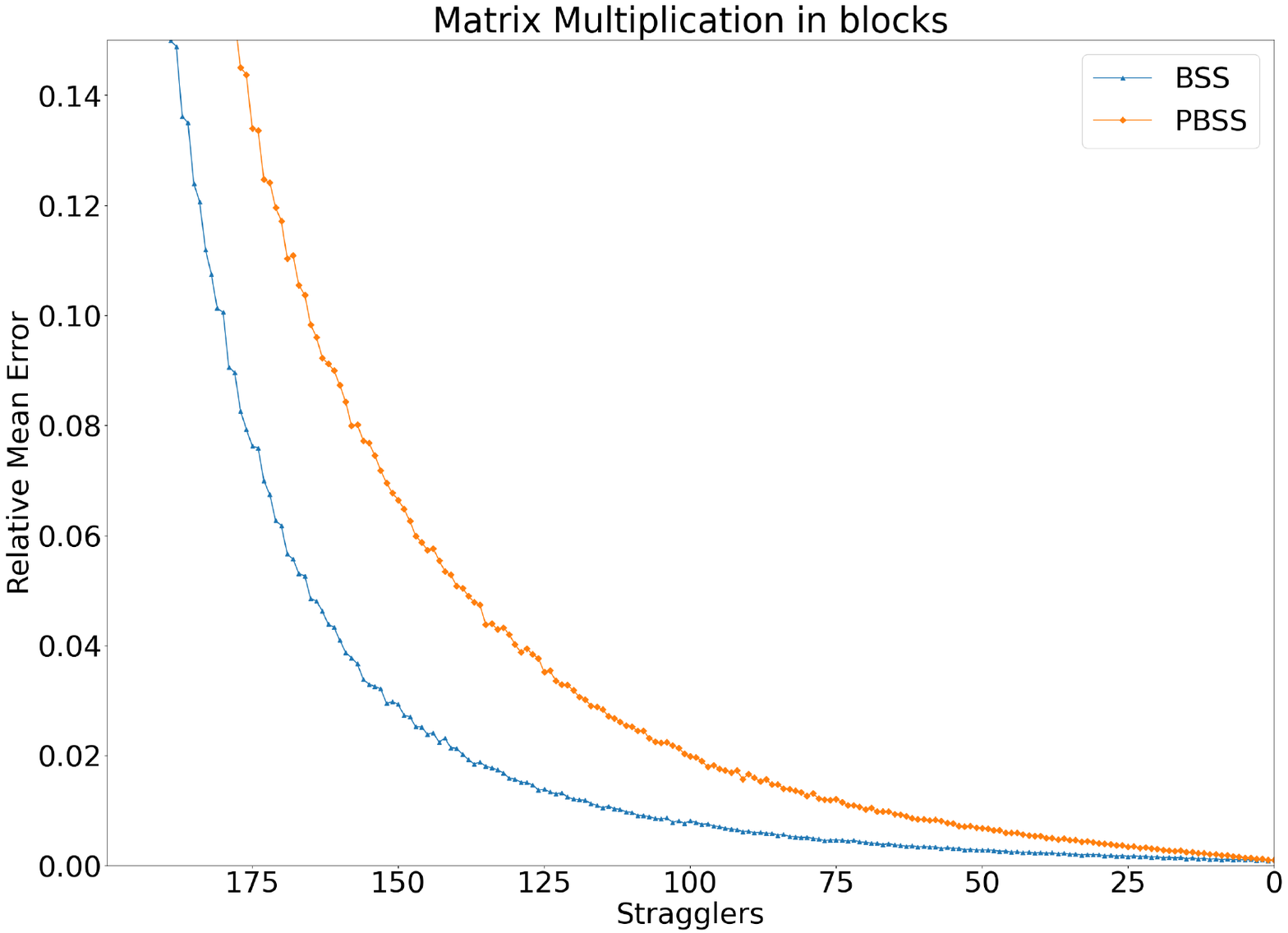}
    }
    \subfloat[Sparse Matrix multiplication divided in 2 vertical blocks.\label{fig:sparse_matrix_mul_blocks_computation_comparison}]{%
        \includegraphics[width=0.50\linewidth]{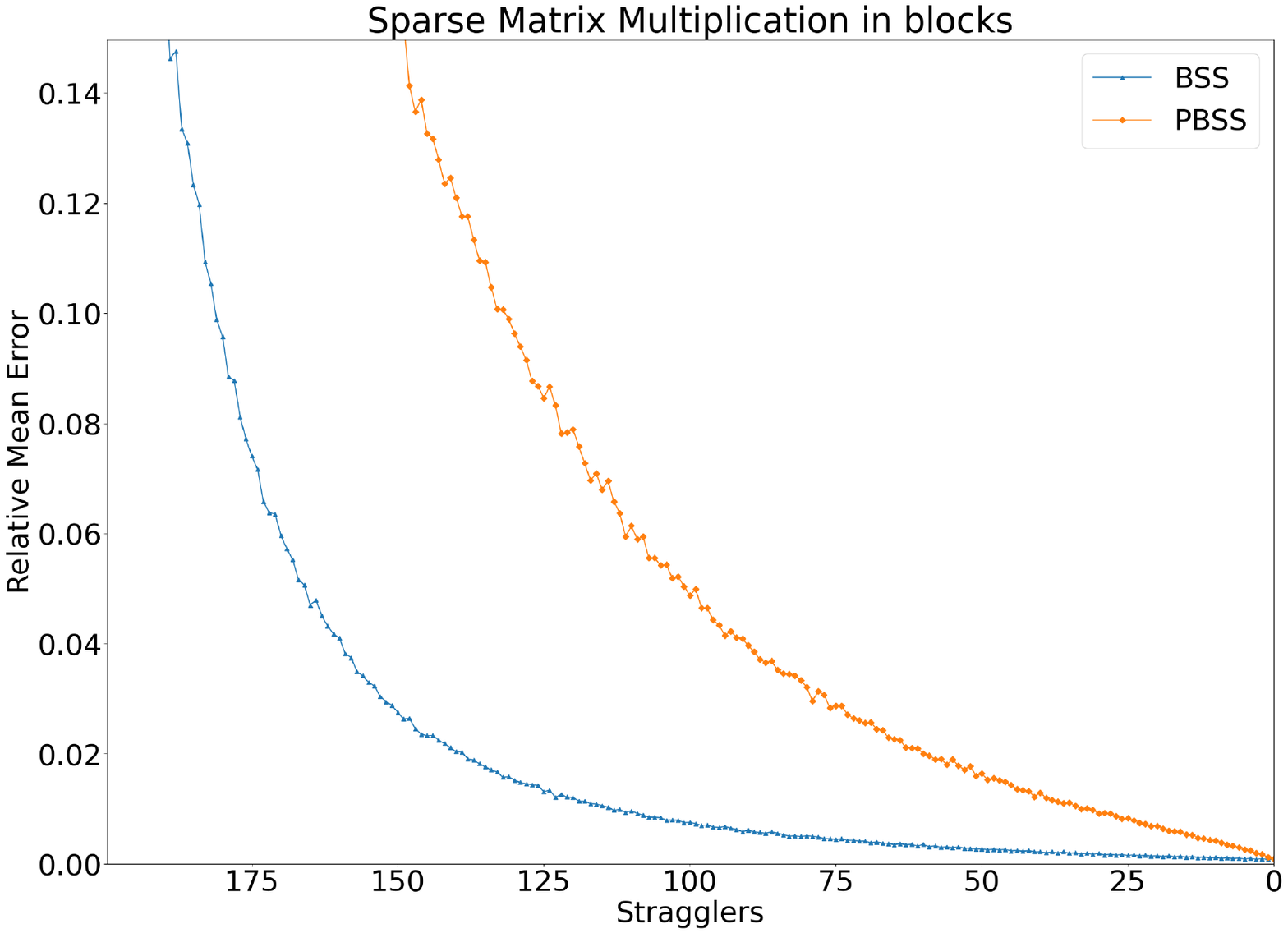}
    }
    \\
    \begin{tabular}{ccc}
    \toprule
    \textbf{Stragglers} & BSS & PBSS \\
    \midrule
    $100$ & $0.008055111$ & $0.019841101$ \\
    $50$ & $0.002768221$ & $0.006725834$ \\
    $0$ & $0.000952427$ & $0.000965228$ \\
    \bottomrule
    \label{tab:matrix_mul_blocks_computation_comparison}
    \end{tabular}
    \begin{tabular}{ccc}
    \toprule
    \textbf{Stragglers} & BSS & PBSS \\
    \midrule
    $100$ & $0.007520580$ & $0.048800873$ \\
    $50$ & $0.002710397$ & $0.016400029$ \\
    $0$ & $0.000928929$ & $0.000949479$ \\
    \bottomrule
    \label{tab:sparse_matrix_mul_blocks_computation_comparison}
    \end{tabular}
    \caption{Matrix Multiplications in blocks. The Tables show the relative errors for private (PBSS) and non-private (BSS) computations.}
    \label{tab:blocks_computation_comparison}
\end{figure*}

Fig.~\ref{tab:blocks_computation_comparison} shows the precision results when applying the matrix multiplication scheme proposed in Section~\ref{sec:matrix_block_division}, where we detailed a way to divide the encoding of a matrix in vertical blocks, in such a way that the multiplication of matrices can be done in smaller blocks. For our analysis, we have used conventional matrices (Fig.~\ref{fig:matrix_mul_blocks_computation_comparison}) and sparse matrices (Fig.~\ref{fig:sparse_matrix_mul_blocks_computation_comparison}. As it can be seen in the tables of results and in the graphs, the numerical precision are identical as the matrix multiplications without dividing the matrices.


Similarly to what happens with the non-linear computations, the matrix products with PBSS show 
precision very close to BSS when the total number of collected results is high. However, matrix
multiplication is specially sensitive to the presence of stragglers. This behaviour is even more
notorious when computing sparse matrices. Thus,  the privacy level in networks with a high 
proportion of straggler nodes must be carefully defined, but it is possible to provide 
strong privacy guarantees in homogeneous networks.

\section{Conclusions}
\label{sec:conclusions}

In this paper we proposed a solution, coined as Private BACC Secret Sharing (PBSS), able to assure privacy in FL and, simultaneously, able to deal with four important aspects that other previous approaches in literature have not solved: (i) working properly with non-linear functions, (ii) working properly with large matrix multiplications, (iii) reduce the high communications overheads due to the presence of semi-hones nodes and (iv) manage stragglers nodes without a high impact on precision. PBSS operate with the model parameters in the coded domain, ensuring output privacy in the nodes without any additional communication cost. Besides, our proposal could be leveraged in secure federated learning for the training of a wide range of machine learning algorithms, since it does not depend on the specific algorithms and it is able to operate with large matrix multiplications.

Since, as we have checked in our analysis, the errors for the aggregation functions are higher than those obtained for activation functions, we are currently working on providing suitable solutions for more complex model aggregation mechanisms in FL. Additionally, we expect to check the security guarantees in more realistic scenarios using well know attacks against our FL proposal.

\bibliographystyle{IEEEtran}
\bibliography{references}  






\end{document}